\documentclass{article}
\usepackage[T1]{fontenc}
\usepackage{iclr2027_conference,times}

\usepackage{amsmath,amsfonts,bm}

\def\eqref#1{equation~\ref{#1}}

\def\1{\bm{1}}

\def\vc{{\bm{c}}}

\def\ve{{\bm{e}}}

\def\vh{{\bm{h}}}

\def\vx{{\bm{x}}}
\def\vy{{\bm{y}}}
\def\vz{{\bm{z}}}

\DeclareMathAlphabet{\mathsfit}{\encodingdefault}{\sfdefault}{m}{sl}
\SetMathAlphabet{\mathsfit}{bold}{\encodingdefault}{\sfdefault}{bx}{n}

\usepackage{amsmath,amssymb}
\usepackage{booktabs}
\usepackage{graphicx}
\usepackage{microtype}
\usepackage{multirow}
\usepackage{makecell}
\usepackage{array}
\usepackage{tabularx}
\newcolumntype{Y}{>{\raggedright\arraybackslash}X}
\usepackage{seqsplit}
\usepackage{float}
\usepackage[section]{placeins}
\usepackage{xcolor}
\usepackage{url}
\usepackage{hyperref}
\hypersetup{
    colorlinks=true,
    linkcolor=black,
    citecolor=black,
    urlcolor=black,
    breaklinks=true,
    pdftitle={PulseBound: Future-Beat State Forecasting Under an Explicit Information Boundary},
    pdfauthor={Chenyang Xu et al.}
}
\usepackage{tikz}
\usetikzlibrary{arrows.meta,positioning,fit,calc}

\definecolor{causalblue}{RGB}{40,92,137}
\definecolor{targetgreen}{RGB}{42,119,92}
\definecolor{auxorange}{RGB}{177,101,32}

\newcommand{\method}{PulseBound}
\newcommand{\code}[1]{\texttt{#1}}

\title{PulseBound: Future-Beat State Forecasting Under an Explicit Information Boundary}

\author{%
  \parbox{0.95\textwidth}{%
    \centering
    {\small
      Chenyang Xu$^{1,3}$, Donglin Xie$^{2}$, Xi Xiang$^{3}$, Xiaoyu Li$^{1}$, Yufan Lu$^{3}$,\\
      Jiqiun Gao$^{3}$, Yi Zhao$^{3}$, Xin-Yi Li$^{1}$, Guangpu Zhu$^{1}$, Zijian Wang$^{1,4}$,\\
      Xiwen Yang$^{1,4}$, Dezhen Wang$^{5}$, Lin Chen$^{6}$, Shenda Hong$^{2}$, Yelei Li$^{1}$}\\[0.35em]
    {\normalfont\mdseries\footnotesize
      \begin{tabular}{@{}l@{}}
      $^{1}$OPPO Health Lab, Guangdong OPPO Mobile Telecommunications Corp., Ltd.\\
      $^{2}$Peking University\\
      $^{3}$Xidian University\\
      $^{4}$University of the Chinese Academy of Sciences\\
      $^{5}$Tongji University\\
      $^{6}$Beijing Technology and Business University
      \end{tabular}}\\[0.25em]
    {\normalfont\mdseries\footnotesize
      \textbf{Correspondence to:} Shenda Hong (\texttt{hongshenda@pku.edu.cn}) and Yelei Li (\texttt{liyelei1@oppo.com}).}
  }%
}

\iclrfinalcopy
\begin{document}
\maketitle
\lhead{Under review}
\begin{abstract}
Predictive representation learning from photoplethysmography (PPG) has a subtle failure mode: causal attention does not guarantee one-sided information access when normalization, nonlocal transforms, or companion views can depend on withheld samples. We introduce \method{}, a PPG representation learner that couples physiologically structured future-beat prediction with an explicit stored-window information boundary. A content-independent cutoff separates a visible prefix from the prediction target; prefix-only normalization, suffix replacement before derived-view construction, and aligned masking make every encoder input a function of the visible prefix and cutoff alone. This yields \emph{stored-suffix invariance}: for fixed model state, randomness, prefix, and cutoff, changing the stored suffix cannot change the forecast context. Within this contract, a shared horizon-conditioned head predicts nine deterministic rhythm and morphology descriptors for up to four extractor-valid future beats, with elementwise validity masks and optional ECG-derived pulse-arrival-time supervision confined to training.On groups held out from \method{} backbone pretraining within MIMIC and VitalDB, \method{} reduces nine-state transformed-space MAE relative to last-visible-beat persistence by 28.06\% and 22.22\%, respectively, and improves MAE, MAE-Skill, and Spearman correlation in all 40 source--cutoff--horizon cells. In a separate seven-model comparison across 13 downstream tasks, \method{} attains the best mean on nine frozen linear-probe tasks and seven full-fine-tuning tasks. Stored-suffix interventions produce zero recorded change in forecast contexts or predictions, with zero suffix-input gradient at audited precision under the declared stored-window interface. Together, these results frame predictive physiological representation learning around three separately testable choices: what information is accessible, what future structure is supervised, and how the learned representation transfers.
\end{abstract}

\section{Introduction}

Photoplethysmography (PPG) is an inexpensive optical measurement of blood-volume dynamics that is ubiquitous in bedside monitors, pulse oximeters, and wearables \citep{allen2007photoplethysmography}. Its availability makes PPG attractive for reusable representation learning across heart and respiratory rate estimation, blood-pressure modeling, stress and activity recognition, and signal-quality assessment. Yet PPG is also highly redundant: adjacent pulse cycles share strong waveform structure, while motion, contact, illumination, and device effects can dominate local variation. Consequently, accurate reconstruction of missing samples need not imply that a representation captures how physiologically meaningful properties evolve.

Recent PPG foundation models address this challenge with complementary supervision. PaPaGei introduces participant- and morphology-aware objectives \citep{pillai2024papagei}; Pulse-PPG emphasizes heterogeneous field recordings \citep{saha2025pulseppg}; AnyPPG uses synchronized ECG for cross-modal guidance \citep{nie2025anyppg}; and GPT-PPG and SIGMA-PPG develop autoregressive and statistically guided generative objectives \citep{chen2025gptppg,guo2026sigmappg}. These approaches motivate two orthogonal design questions: \emph{what should a representation predict, and exactly what observations may it use to make that prediction?}

The second question is easy to under-specify. Masking future tokens or using causal attention does not prevent hidden samples from influencing visible inputs through full-window normalization, zero-phase or global transforms, or an unmasked companion view. Likewise, selecting a cutoff from future beat detections can leak target-dependent timing. For forecasting, the relevant object is therefore not the attention mask alone, but the complete computation that constructs the forecast context from admissible observations.

We introduce \method{}, which makes this conditioning set explicit. From a stored, preprocessed 30-s PPG window, we sample a patch-aligned cutoff independently of waveform content, metadata, beat detections, and target values. Normalization uses visible-prefix statistics; hidden samples are removed before nonlocal derived views are formed; and aligned hidden tokens are masked across streams. The resulting encoder inputs depend only on the stored prefix and cutoff. This yields \emph{stored-suffix invariance}: with model state, randomness, prefix, and cutoff fixed, replacing the stored suffix cannot change the forecast context. The guarantee begins at the stored-window interface; it is intentionally narrower than end-to-end streaming causality, which would also require boundary-preserving upstream preprocessing.

Within this boundary, \method{} predicts interpretable future-beat states rather than requiring sample-wise waveform reproduction. A shared horizon-conditioned head forecasts nine deterministic PPG-derived descriptors spanning inter-beat timing, amplitude, pulse morphology, baseline, and multiscale energy for up to four extractor-valid future beats. Component-wise validity masks exclude inadmissible targets. When synchronized ECG is available, pulse arrival time (PAT) provides an additional training-only supervision signal, while ECG never enters the encoder or deployment path. The resulting targets are inspectable signal-derived proxies rather than clinical ground truth.

Our contributions are threefold:
\begin{itemize}
\item \textbf{An explicit, auditable information boundary.} We formalize cutoff-relative stored-suffix invariance and give a construction based on content-independent cutoff sampling, prefix-only normalization, suffix replacement before derived transforms, and aligned masking. The guarantee concerns model-side access from the stored preprocessed window onward and is distinct from attention causality or raw-signal streaming equivalence.

\item \textbf{Validity-aware prediction of interpretable future beats.} Under this contract, a shared multi-horizon predictor estimates nine deterministic PPG-derived rhythm and morphology descriptors and optional training-only ECG-derived PAT for up to four future beats, while component-wise masks prevent invalid measurements from being treated as zeros.

\item \textbf{Separate tests of conditional forecasting and transferable representation quality.} We evaluate future-state prediction against train-median and last-visible-beat references on identical paired support, audit the declared information boundary by intervention and input gradients, and independently compare the complete pretrained system across 13 downstream tasks. Checkpoint-compatible frontends are retained for released baselines, so downstream comparisons are interpreted at the system level rather than as causal attribution to a single pretraining objective.
\end{itemize}

\section{Related Work}
\label{sec:related}

\paragraph{PPG representation learning and predictive self-supervision.}
PPG representation learning has explored participant structure, pulse morphology,
acquisition diversity, and predictive or generative supervision.
PaPaGei uses participant- and morphology-aware objectives
\citep{pillai2024papagei}; Pulse-PPG emphasizes heterogeneous field recordings
\citep{saha2025pulseppg}; DNA-PPG aligns morphology-aware neighborhoods with
continuous physiological semantics \citep{yang2026dnappg}; and large-scale
pulse-oximetry pretraining further demonstrates transferable physiological
structure in optical waveforms \citep{kohn2026pulseoxfm}.
Predictive and generative approaches include GPT-PPG, SIGMA-PPG, TS2TC, and
RhythmJEPA
\citep{chen2025gptppg,guo2026sigmappg,zhang2024ts2tc,nguyen2026rhythm},
alongside broader masked, contrastive, and predictive representation learning
\citep{he2022mae,oord2018cpc,eldele2021tstcc,yue2022ts2vec}.
A recent ECG/PPG review organizes physiological SSL along related
reconstruction, contrastive, predictive, and cross-modal axes
\citep{ding2024sslreview}.
Our focus is complementary to the pretext objective itself: \method{} asks not
only \emph{what} should be predicted, but also \emph{which stored observations
may influence the prediction context}.

\paragraph{Blood-pressure transfer, subject specificity, and signal quality.}
Prior work also exposes representation properties directly relevant to our
evaluation.
UPR-BP studies unsupervised PPG representation learning for noninvasive
blood-pressure estimation \citep{ma2024uprbp}, while STP develops
self-supervised transfer for PPG-based BP prediction \citep{ma2024stp}.
These works establish BP-relevant self-supervised transfer; we therefore do not
claim that BP information is unique to \method{}, and avoid numerical rankings
across unmatched BP-specific protocols.
Separately, Ghorbani et al.\ report substantial inter-subject structure in
self-supervised PPG representations \citep{ghorbani2023intersubject}, while
work on quasiperiodic time series highlights a related tension between
record-specific structure and within-record temporal variation
\citep{atienza2024quasiperiodic}.
This is pertinent to our subject-contrastive objective and closed-set
identification task: subject-typed registers are architectural roles, not
evidence of subject-invariant physiology.
Signal quality is likewise a distinct representation problem.
Shao et al.\ study self-supervised, device-agnostic PPG quality assessment
\citep{shao2025ppgquality}; our quality scaffold and Stanford evaluation instead
measure quality-sensitive transfer within a multi-task representation and do
not establish device invariance.
More generally, deterministic PPG targets remain dependent on beat detection
and signal quality \citep{charlton2022detecting}.

\paragraph{Cross-modal physiological supervision.}
Paired biosignals can enrich representation learning without increasing
deployment-time sensing requirements.
AnyPPG uses ECG-guided PPG representation learning
\citep{nie2025anyppg}, and ECG-guided heart-age modeling similarly transfers
cross-modal supervision to PPG-only inference \citep{xie2026heartage}.
xMAE exploits ECG--PPG structure for masked cross-modal reconstruction
\citep{zhou2026crossmodal}; CardioPPG combines cross-modal representation
learning with ECG synthesis \citep{ding2026cardioppg}; and CardioFM jointly
models ECG and PPG with multimodal attention
\citep{hassanuzzaman2026cardiofm}.
In \method{}, ECG is restricted to training-time supervision: neither ECG
waveforms nor ECG-derived labels enter the encoder or forecast context, and
deployment remains PPG-only.

\paragraph{Positioning of \method{}.}
Morphology-aware supervision, BP-oriented transfer, subject-sensitive
representations, quality modeling, predictive pretraining, and cross-modal
scaffolds are therefore all present in prior work.
The distinction of \method{} is their coupling to an explicit one-sided
information-access contract.
Content-independent cutoff sampling, prefix-only normalization, suffix
replacement before nonlocal derived transforms, and aligned masking make every
encoder input a function of the stored prefix.
Once this boundary is established, attention, registers, auxiliary objectives,
and future-beat targets determine what is learned \emph{within} the admissible
information set rather than which information is available.
The fixed descriptor targets make the supervision inspectable, while the
boundary makes its conditioning path auditable.
Our downstream benchmark consequently characterizes system-level transfer with
checkpoint-compatible frontends rather than attributing gains causally to any
single objective.

\section{Method}
\label{sec:method}

\begin{figure}[t]
    \centering
    \includegraphics[width=\linewidth]{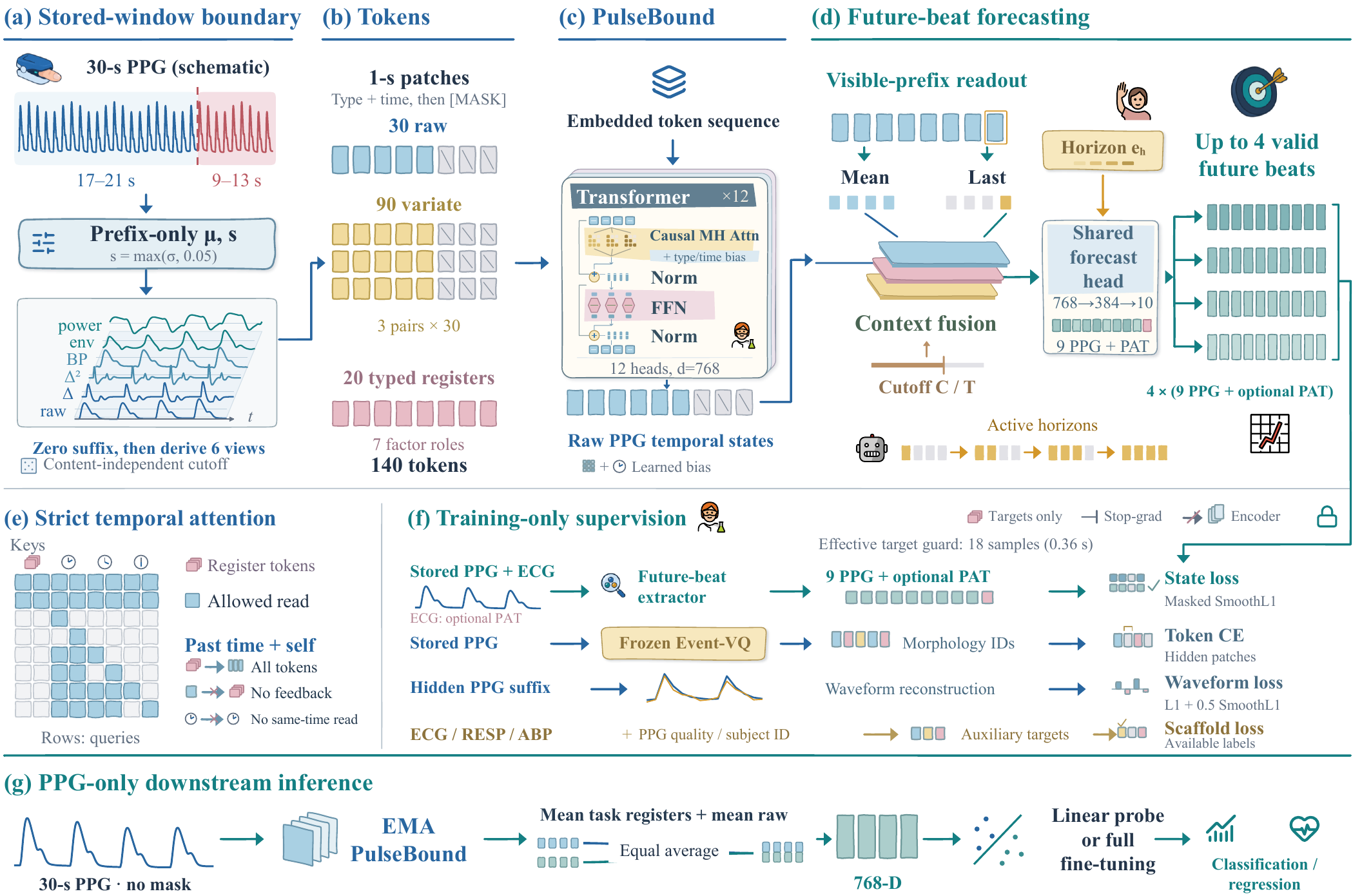}
    \caption{\method{} separates prefix-conditioned encoding from training-only supervision while preserving PPG-only inference.}
    \label{fig:framework}
\end{figure}

\method{} separates the \emph{conditioning path} from the \emph{supervision path}. The encoder receives only prefix-dependent inputs, while a training-only extractor may inspect the stored waveform and optional synchronized ECG to construct prediction targets. Future-state prediction is optimized jointly with masked-token, waveform, and auxiliary objectives on the shared backbone, but none of those target signals enters the forecast context (Figure~\ref{fig:framework}).

\subsection{Forecasting task and information-boundary contract}

Let $\vx_b\in\mathbb{R}^{L}$ be a stored, preprocessed 30-s PPG window sampled at $f_s=50$~Hz, with $L=1500$. Patches of $P=50$ samples yield $T=30$ temporal positions. Given mask-ratio bounds $(r_{\min},r_{\max})=(0.30,0.45)$ and minimum context/future lengths $(T_{\mathrm{ctx}}^{\min},T_{\mathrm{fut}}^{\min})=(12,8)$, the admissible hidden-patch counts satisfy
\begin{equation}
 H_{\min}=\max\!\left(T_{\mathrm{fut}}^{\min},\lceil r_{\min}T\rceil\right),\quad
 H_{\max}=\min\!\left(T-T_{\mathrm{ctx}}^{\min},\lfloor r_{\max}T\rfloor\right).
 \label{eq:hidden-range}
\end{equation}
We sample $H_b\sim\mathrm{Uniform}\{H_{\min},\ldots,H_{\max}\}$ independently of waveforms, metadata, beat detections, and target scores, set $C_b=T-H_b$, and define
\begin{equation}
 M_{b,t}=\mathbf{1}[t\ge C_b],\qquad t\in\{0,\ldots,T-1\}.
 \label{eq:mask}
\end{equation}
Here $H_b$ counts hidden patches, whereas $h$ indexes future beats. The configuration gives $(H_{\min},H_{\max})=(9,13)$ and cutoff time $C_bP/f_s\in\{17,\ldots,21\}$~s. Unlike a boundary chosen using future beat detections, this sampling rule cannot encode target-dependent event timing.

The task is to predict up to four extractor-valid future beats from the prefix. With model parameters and persistent state fixed, let $\xi$ denote waveform-independent model randomness and define
\begin{equation}
 \mathcal{I}^{\xi}_b=\sigma(\vx_{b,<C_bP},C_b,\xi).
 \label{eq:information-set}
\end{equation}
A forecast context obeys the contract when it is measurable with respect to $\mathcal{I}^{\xi}_b$. Labels and validity masks may depend on the stored suffix, but cannot enter the encoder or forecast context. This is a forward-access constraint beginning at the stored-window interface, not a guarantee about upstream raw-signal preprocessing or statistical independence between past and future physiology.

\subsection{Boundary-preserving signal and token construction}

For $N_b=C_bP$ visible samples, normalization uses prefix population statistics,
\begin{equation}
 \mu_b=\frac{1}{N_b}\sum_{i<N_b}x_{b,i},\qquad
 s_b=\max\!\left(\sqrt{\frac{1}{N_b}\sum_{i<N_b}(x_{b,i}-\mu_b)^2},\tau\right),
 \quad \tau=0.05.
 \label{eq:prefixnorm}
\end{equation}
The scale floor is fixed using the MIMIC--VitalDB training split and stabilizes low-variance prefixes without consulting the current suffix. We standardize the full array with $(\mu_b,s_b)$, then construct two token streams.

For the \emph{raw stream}, each patch is independently projected to $d=768$ dimensions, and every hidden token is overwritten by a learned mask token. Patch-local projection prevents hidden samples from affecting visible raw tokens. For the \emph{derived streams}, hidden standardized samples are first replaced by zero. Only then do we compute six views: normalized raw PPG, first and second finite differences, a 0.5--8~Hz zero-phase FFT band-pass signal, a Hilbert amplitude envelope, and a smoothed log band-power envelope. Subsequent view-wise z-scoring therefore remains suffix-independent. Ordered by physiological type, the views are paired and embedded into three aligned variate tokens per position, with type and sinusoidal time embeddings; hidden variate tokens are also replaced by mask tokens. Together with 20 learned registers, this yields $30+3\times30+20=140$ tokens.

These views are \emph{prefix-global}: an earlier visible feature may depend on later samples within the same prefix. They preserve the cutoff-relative information boundary without implying sample-wise streaming causality.

\subsection{Temporal connectivity and global readouts}

The $d=768$, 12-layer backbone uses 20 typed registers. Temporal queries may read only earlier temporal tokens and themselves, while registers aggregate the masked sequence without feeding information back into temporal states. Learned type-pair and relative-time biases operate only within this routing; their exact parameterization is given in Appendix~\ref{app:architecture}. This connectivity controls representation mixing \emph{after} prefix-dependent inputs have been constructed and is therefore separate from the information-boundary guarantee.

\paragraph{Proposition 1 (stored-window suffix invariance).}

Fix model parameters, persistent state, and randomness $\xi$. For any two stored windows sharing the same prefix and cutoff, with suffixes for which the forward computation is well defined,
\begin{equation}
 \vc_{\theta}([\vx_{<CP};\vx^{s}],C;\xi)
 =\vc_{\theta}([\vx_{<CP};\tilde{\vx}^{s}],C;\xi).
 \label{eq:suffix-invariance}
\end{equation}
\emph{Proof.} Let $U_{\theta,C}(\vx)$ collect all encoder inputs, including tokens, registers, embeddings, and masks. Prefix-only normalization, patch-local raw projection followed by masking, and suffix replacement before derived transforms ensure
$U_{\theta,C}(\vx)=G_{\theta}(\vx_{<CP},C)$, independent of the stored suffix. The forecast context can therefore be written as
$\vc_{\theta}(\vx,C;\xi)=F_{\theta}(U_{\theta,C}(\vx),C;\xi)$.
Equal prefixes thus yield identical contexts, although their targets may differ, proving Equation~\ref{eq:suffix-invariance}. The guarantee follows from input construction rather than temporal routing. It applies from the stored, preprocessed-window interface onward and does not cover upstream preprocessing.
\subsection{Deterministic future-beat supervision}

A training-only deterministic extractor identifies up to four future beats from the stored waveform using prefix-based normalization, a 0.5--8~Hz filtered PPG, refractory peak detection, and an effective 18-sample (0.36-s) future-peak guard. For each accepted beat,
\begin{equation}
 \vy_{b,h}=\mathcal{T}\!\left(
 \mathrm{IBI},A,R,W_{1/2},S,B,Q,E_{\mathrm{fast}},E_{\mathrm{mid}},\mathrm{PAT}
 \right)\in\mathbb{R}^{10},
 \label{eq:state}
\end{equation}
where the first nine components are PPG-derived descriptors of timing, amplitude, morphology, baseline, and multiscale energy, and PAT is optional. Horizons index accepted events rather than fixed elapsed times. An elementwise validity mask excludes inadmissible measurements without treating them as zeros; PAT validity affects only the PAT component. Neither synchronized ECG nor extracted targets enter the encoder or forecast head. Appendix~\ref{app:targets} gives the exact transforms, detection rules, and validity conditions.

\subsection{Horizon-conditioned forecast and curriculum}

Let $\vh_b^{\mathrm{last}}$ be the raw PPG state at $C_b-1$, and $\bar{\vh}_b$ the mean over raw states with $t<C_b$. A shared predictor computes
\begin{align}
 \vc_b &= f_c([\vh_b^{\mathrm{last}};\bar{\vh}_b])+f_u(C_b/T),\\
 \hat{\vy}_{b,h} &= f_p(\vc_b+\ve_h),\qquad h\in\{1,2,3,4\},
 \label{eq:head}
\end{align}
where $\ve_h$ selects the future event and $C_b/T$ specifies the known context length. The fusion $f_c$ is LayerNorm--Linear--SiLU--Dropout from $2d$ to $d$; $f_u$ maps the cutoff fraction to $d$; and $f_p$ is a LayerNorm MLP with widths $d\rightarrow d/2\rightarrow10$. Dropout is 0.1. All horizons share $f_p$.

During training, a four-phase curriculum progressively activates horizons $H_1$--$H_4$, while a masked smooth-$L_1$ objective averages over valid active state elements. Input sampling, cutoff policy, and target definitions remain fixed throughout the curriculum. The exact schedule and reduction are given in Appendix~\ref{app:exact-objective}.

\subsection{Joint optimization and PPG-only inference}

Future-state prediction is optimized jointly with masked Event-VQ token prediction, hidden-waveform reconstruction, available ECG/respiration/blood-pressure/quality scaffold losses, subject contrastive supervision, and the ponder regularizer. Exact objective definitions and weights are given in Appendix~\ref{app:exact-objective}; tokenizer and scaffold details appear in Appendix~\ref{app:architecture}, and optimization settings in Appendix~\ref{app:repro}. No teacher network is used, and all 12 backbone layers execute during both training and inference.

Downstream inference encodes a complete 30-s PPG window without masking, a tokenizer, or auxiliary modalities. The exported representation is
\begin{equation}
 \vz=\tfrac12\left(\operatorname{mean}(\text{task registers})+
 \operatorname{mean}(\text{PPG temporal tokens})\right).
 \label{eq:downstream-representation}
\end{equation}
This 768-dimensional full-window transfer readout combines global and temporal features and is distinct from the visible-prefix context used for forecasting.
\section{Experiments}
\label{sec:experiments}

We evaluate downstream transfer and future-beat forecasting under the stored-prefix contract. Complete-system and downstream-ablation results aggregate three pretraining seeds, while the main component-wise forecasting ablation uses seed 0; separate three-seed forecasting controls are reported on available-case support. Additional horizon, cutoff, ECG-conditional, and boundary analyses characterize robustness and scope.

\subsection{Evaluation protocol}
\label{sec:exp-protocol}

\paragraph{Pretraining and model selection.}
We pretrain the \method{} backbone for 80 epochs on the MIMIC-III Waveform Database Matched Subset \citep{moody2020mimic} and VitalDB \citep{lee2022vitaldb}, using 90/5/5 patient-group splits, an 80:20 per-epoch source mixture, and seeds $\{0,1,2\}$. Downstream transfer uses zero-based epoch 29, selected by validation-only linear probing without test access, whereas future-state evaluation and the stored-suffix audit use terminal epoch 79 without test-based selection. The frozen Event-VQ tokenizer predates the MIMIC split and was trained on the full processed MIMIC corpus; thus MIMIC group-disjoint claims apply specifically to backbone pretraining. The information-access guarantee begins at the stored, preprocessed-window interface. Full training, tokenizer, checkpoint-selection, and preprocessing details appear in Appendices~\ref{app:architecture},~\ref{app:final-eval-protocol}, and~\ref{app:repro}.

\paragraph{Downstream and intrinsic evaluation.}
We evaluate transfer on 13 tasks spanning BIDMC \citep{pimentel2017bidmc}, PPG-BP \citep{liang2018ppgbp}, WESAD \citep{schmidt2018wesad}, DaLiA \citep{reiss2019deepppg}, Real-World PPG \citep{siam2019realworld}, and Stanford \citep{torressoto2020stanford}. Frozen linear probing is the primary transfer setting, while full fine-tuning is reported as a secondary end-to-end adaptation test. All methods use matched evaluation examples, labels, outer splits, and task metrics, while retaining checkpoint-compatible frontends. Intrinsic forecasting is evaluated separately on group-disjoint held-out partitions using identical paired support for \method{} and reference predictors. Exact task definitions, metrics, split construction, seed aggregation, and uncertainty conventions are given in Appendix~\ref{app:final-eval-protocol}.

\begin{table}[!htbp]
\caption{Cross-model downstream performance across 13 PPG tasks under frozen linear probing.}
\label{tab:cross-model-lp}
\centering
\scriptsize
\setlength{\tabcolsep}{1.2pt}
\renewcommand{\arraystretch}{1.12}

\begin{tabular*}{\linewidth}{@{\extracolsep{\fill}}lccccccc@{}}
\multicolumn{8}{c}{\textit{Regression tasks (MAE $\downarrow$)}}\\[-1.5pt]
\toprule
& \multicolumn{3}{c}{BIDMC} & \multicolumn{3}{c}{PPG-BP} & \\
\cmidrule(lr){2-4}\cmidrule(lr){5-7}\cmidrule(lr){8-8}
Method & HR & RR & SpO$_2$ & SBP & DBP & HR & Avg. \\
\midrule
PaPaGei-S
& 5.688$\pm$0.103 & 2.182$\pm$0.002 & 2.483$\pm$0.011
& 16.425$\pm$0.019 & 8.744$\pm$0.055 & 8.615$\pm$0.041 & 7.356 \\
PaPaGei-P
& 6.582$\pm$0.049 & 2.274$\pm$0.044 & \textbf{2.411$\pm$0.015}
& 16.497$\pm$0.080 & 8.812$\pm$0.087 & 8.533$\pm$0.062 & 7.518 \\
PaPaGei-S-SVRI
& 7.072$\pm$0.028 & \textbf{2.145$\pm$0.009} & 2.475$\pm$0.016
& 16.471$\pm$0.043 & 8.752$\pm$0.013 & 8.581$\pm$0.018 & 7.583 \\
AnyPPG
& 4.906$\pm$0.577 & 2.492$\pm$0.092 & 2.976$\pm$0.401
& 16.852$\pm$0.113 & 9.038$\pm$0.045 & 7.683$\pm$0.214 & 7.324 \\
Pulse-PPG
& 4.061$\pm$0.194 & 2.890$\pm$0.166 & 3.278$\pm$0.240
& 19.812$\pm$0.302 & 11.557$\pm$0.536 & 10.871$\pm$0.138 & 8.745 \\
SIGMA-PPG
& 3.730$\pm$0.064 & 2.247$\pm$0.050 & 2.531$\pm$0.008
& 16.450$\pm$0.186 & 8.813$\pm$0.112 & 7.123$\pm$0.088 & 6.816 \\
\midrule
\textbf{\method{}}
& \textbf{1.974$\pm$0.559} & 2.263$\pm$0.042 & 2.521$\pm$0.050
& \textbf{14.913$\pm$0.295} & \textbf{8.554$\pm$0.036} & \textbf{5.288$\pm$0.965} & \textbf{5.919} \\
\bottomrule
\end{tabular*}

\vspace{5pt}

\begin{tabular*}{\linewidth}{@{\extracolsep{\fill}}lcccccccc@{}}
\multicolumn{9}{c}{\textit{Classification and recognition tasks ($\uparrow$)}}\\[-1.5pt]
\toprule
& PPG-BP & \multicolumn{3}{c}{WESAD} & DaLiA & Real-World & Stanford & \\
\cmidrule(lr){2-2}\cmidrule(lr){3-5}\cmidrule(lr){6-6}
\cmidrule(lr){7-7}\cmidrule(lr){8-8}\cmidrule(lr){9-9}
Method
& \makecell{HTN\\AUC}
& \makecell{Stress\\AUC}
& \makecell{Affect\\F1}
& \makecell{Emotion\\AUC}
& \makecell{Act.\\F1}
& \makecell{ID\\F1}
& \makecell{Qual.\\mAUC} & Avg. \\
\midrule
PaPaGei-S
& 0.495$\pm$0.009 & 0.834$\pm$0.001 & 0.270$\pm$0.011
& \textbf{0.525$\pm$0.021} & 0.116$\pm$0.010 & 0.047$\pm$0.002
& 0.874$\pm$0.001 & 0.452 \\
PaPaGei-P
& 0.479$\pm$0.028 & 0.874$\pm$0.004 & 0.384$\pm$0.008
& 0.522$\pm$0.063 & 0.240$\pm$0.005 & 0.471$\pm$0.008
& 0.822$\pm$0.002 & 0.542 \\
PaPaGei-S-SVRI
& 0.521$\pm$0.030 & 0.808$\pm$0.004 & 0.365$\pm$0.009
& 0.486$\pm$0.020 & 0.121$\pm$0.002 & 0.138$\pm$0.002
& 0.852$\pm$0.003 & 0.470 \\
AnyPPG
& 0.516$\pm$0.032 & 0.918$\pm$0.003 & 0.488$\pm$0.011
& 0.417$\pm$0.037 & 0.249$\pm$0.005 & 0.631$\pm$0.008
& 0.911$\pm$0.003 & 0.590 \\
Pulse-PPG
& 0.517$\pm$0.009 & 0.821$\pm$0.002 & 0.409$\pm$0.003
& 0.495$\pm$0.034 & 0.214$\pm$0.004 & \textbf{0.755$\pm$0.005}
& 0.900$\pm$0.001 & 0.587 \\
SIGMA-PPG
& 0.475$\pm$0.018 & 0.839$\pm$0.005 & 0.443$\pm$0.002
& 0.483$\pm$0.065 & 0.189$\pm$0.005 & 0.309$\pm$0.007
& 0.886$\pm$0.004 & 0.518 \\
\midrule
\textbf{\method{}}
& \textbf{0.617$\pm$0.024} & \textbf{0.953$\pm$0.007}
& \textbf{0.563$\pm$0.025} & 0.496$\pm$0.073
& \textbf{0.290$\pm$0.008} & 0.698$\pm$0.007
& \textbf{0.934$\pm$0.005} & \textbf{0.650} \\
\bottomrule
\end{tabular*}

\vspace{2pt}
\parbox{\linewidth}{\scriptsize
\textit{Aggregation.}
For \method{}, mean$\pm$sample SD is across the three pretraining seeds
after averaging downstream seeds within each pretraining seed; released
baselines use downstream seeds 42/43/44 directly. BIDMC is shown
task-by-task but excluded from the no-overlap aggregate when pretraining
overlap cannot be ruled out. The final column is the arithmetic mean of the task means in the panel. Because it combines heterogeneous tasks and, for regression, different physical units, it is reported only as a descriptive summary and carries no cross-task uncertainty estimate.
}
\end{table}

\subsection{Downstream transfer and ablation analysis}
\label{sec:exp-transfer-ablation}

\paragraph{Cross-model comparison.}
Frozen linear probing is our primary transfer test. Table~\ref{tab:cross-model-lp} compares \method{} with three PaPaGei variants (P, S, and S-SVRI), AnyPPG, Pulse-PPG, and SIGMA-PPG; full fine-tuning is reported separately in Appendix Table~\ref{tab:cross-model-ft}. \method{} attains the best mean on 9/13 LP tasks, spanning all four PPG-BP tasks, WESAD stress and affect, DaLiA activity, Stanford quality, and the BIDMC HR task. The pattern is not uniform: other systems lead BIDMC RR/SpO$_2$, WESAD emotion, and Real-World identification. Under full fine-tuning, \method{} attains the best mean on 7/13 tasks. BIDMC is retained task-by-task but excluded from overlap-sensitive summaries because pretraining overlap cannot be ruled out. These results compare complete pretrained systems; they do not isolate the causal contribution of any single objective.

\paragraph{State-supervision ablation.}
We use two diagnostic controls to separate target alignment from generic predictor capacity. Forecasting ablations use pretraining seed 0 at zero-based epoch 79. The no-state control removes future-state supervision from pretraining and fits a capacity-matched MLP to the frozen representation; the shuffled-target control retains the forecasting head but destroys alignment between examples and future-state targets.

\begin{table}[!htbp]
\caption{Core seed-0 state-supervision ablation at zero-based epoch 79. Forecast Skill equally aggregates nine PPG states, five cutoffs, four horizons, and two sources relative to the training-median reference. Complete ablations and three-seed available-case controls appear in Appendix Tables~\ref{tab:ablation-main} and~\ref{tab:forecast-controls-three-seed}.}
\label{tab:state-ablation-mini}
\centering
\small
\setlength{\tabcolsep}{8pt}
\begin{tabular}{@{}lr@{}}
\toprule
Condition & Future-state macro Skill $\uparrow$ \\
\midrule
Full \method{} & 0.3926 \\
No-state control & 0.3527 \\
Shuffled-target control & 0.0032 \\
\bottomrule
\end{tabular}
\end{table}

At this fixed seed-0 evaluation, full \method{} exceeds the capacity-matched no-state predictor by 0.0399 Skill, while the shuffled-target direct predictor is near the training-median reference (0.0032 Skill). These controls support the importance of aligned state supervision for the evaluated forecasting head, but they do not imply that the no-state representation contains no recoverable future information. The complete multi-seed downstream ablation, including PAT, curriculum, view, and source controls, is reported in Appendix~\ref{app:complete-ablation}; a separate three-seed forecasting-control analysis is reported in Appendix Table~\ref{tab:forecast-controls-three-seed}.

\subsection{Future-beat state forecasting}
\label{sec:exp-future-state}

\paragraph{Test support.}
We evaluate MIMIC's 1,000,765 locked-test windows from 314 groups and VitalDB's 58,189 windows from 294 groups. Both sources participate in backbone pretraining; these tests measure group-disjoint generalization with respect to the \method{} backbone splits, not unseen-source generalization. The frozen Event-VQ tokenizer was trained on the full processed MIMIC corpus, as documented in Appendix~\ref{app:architecture}. All five cutoffs $t_C=CP/f_s\in\{17,18,19,20,21\}$~s are evaluated. Horizons $H_1$--$H_4$ index the first four extractor-valid future beats after the 18-sample (0.36-s) guard, rather than fixed elapsed times. The primary outcome covers the nine PPG-derived states in Equation~\ref{eq:state}; ECG-conditional PAT is separate. Invalid elements are masked, not scored as zeros. Each source--cutoff--horizon--state cell requires at least 100 valid windows from 20 groups, with target, persistence, and model predictions restricted to identical paired support.

\paragraph{Metrics and references.}
We use two references: a training-set median specific to source, cutoff, horizon, and state, and last-visible-beat persistence. Seed-0 objective and source comparisons are reported separately in Table~\ref{tab:ablation-main}. Let $m^{\mathrm{train}}_{s,C,h,k}$ be the median fitted using source $s$'s training split, and $\mathcal{S}_{s,C,h,k}$ the shared test support. We define
\begin{equation}
 \operatorname{MAE\text{-}Skill}_{s,C,h,k}
 =1-\frac{
 \operatorname{MAE}_{\mathcal{S}_{s,C,h,k}}(\hat y,y)}
 {\operatorname{MAE}_{\mathcal{S}_{s,C,h,k}}
 (m^{\mathrm{train}}_{s,C,h,k},y)}.
 \label{eq:mae-skill-exp}
\end{equation}
The reference value is training-fitted; both errors are evaluated on the same test support with identical weights. MAE uses patient-macro aggregation within each state. Skill is computed state-wise before averaging the nine states equally; it is not reconstructed from a ratio of macro-averaged MAEs. Spearman's $\rho$ pools valid windows within each state and then averages correlations equally across states. Cutoffs, and horizons where indicated, are averaged within each checkpoint before seed aggregation. Coverage describes extractor-defined support under the pairing rules, not agreement with independent physiological annotations.

\begin{table}[!htbp]
\centering
\caption{Future-beat forecasting on identical paired support. MAE and MAE-Skill use patient-macro aggregation within each of nine PPG-derived states, followed by equal state weighting; $\rho$ averages state-wise micro Spearman correlations. Values average five cutoffs within each checkpoint. \method{} reports mean$\pm$sample SD over three seeds; reference predictors are deterministic. Train-median Skill is zero and its within-cell correlation is undefined. Coverage describes paired support, not target correctness.}
\label{tab:future-macro}
\scriptsize
\setlength{\tabcolsep}{2.8pt}
\resizebox{\textwidth}{!}{%
\begin{tabular}{@{}llcccccccc@{}}
\toprule
& & & \multicolumn{3}{c}{Persistence} &
\multicolumn{3}{c}{\method{}} & \\
\cmidrule(lr){4-6}\cmidrule(lr){7-9}
Source & Horizon & Train-median MAE & MAE $\downarrow$ & Skill $\uparrow$ &
$\rho\uparrow$ & MAE $\downarrow$ & Skill $\uparrow$ & $\rho\uparrow$ &
Coverage (\%) \\
\midrule
\multirow{4}{*}{MIMIC}
& $H_1$ & 0.16536 & 0.14040 & 0.1180 & 0.5083 &
$\mathbf{0.09814\pm0.00017}$ & $\mathbf{0.3899\pm0.0011}$ &
$\mathbf{0.7127\pm0.0008}$ & 99.64 \\
& $H_2$ & 0.16534 & 0.14514 & 0.0879 & 0.4814 &
$\mathbf{0.10458\pm0.00006}$ & $\mathbf{0.3480\pm0.0004}$ &
$\mathbf{0.6693\pm0.0005}$ & 99.67 \\
& $H_3$ & 0.16480 & 0.14585 & 0.0807 & 0.4818 &
$\mathbf{0.10604\pm0.00006}$ & $\mathbf{0.3365\pm0.0004}$ &
$\mathbf{0.6596\pm0.0002}$ & 99.67 \\
& $H_4$ & 0.16470 & 0.14698 & 0.0731 & 0.4751 &
$\mathbf{0.10731\pm0.00006}$ & $\mathbf{0.3277\pm0.0004}$ &
$\mathbf{0.6496\pm0.0002}$ & 99.66 \\
\midrule
\multirow{4}{*}{VitalDB}
& $H_1$ & 0.14890 & 0.10428 & 0.2830 & 0.6469 &
$\mathbf{0.07869\pm0.00015}$ & $\mathbf{0.4663\pm0.0010}$ &
$\mathbf{0.7777\pm0.0006}$ & 99.23 \\
& $H_2$ & 0.15000 & 0.10763 & 0.2620 & 0.6300 &
$\mathbf{0.08341\pm0.00012}$ & $\mathbf{0.4341\pm0.0009}$ &
$\mathbf{0.7519\pm0.0007}$ & 99.24 \\
& $H_3$ & 0.15013 & 0.10755 & 0.2631 & 0.6298 &
$\mathbf{0.08479\pm0.00004}$ & $\mathbf{0.4242\pm0.0003}$ &
$\mathbf{0.7432\pm0.0003}$ & 99.16 \\
& $H_4$ & 0.15013 & 0.10811 & 0.2595 & 0.6266 &
$\mathbf{0.08566\pm0.00005}$ & $\mathbf{0.4177\pm0.0004}$ &
$\mathbf{0.7381\pm0.0002}$ & 99.09 \\
\bottomrule
\end{tabular}%
}
\end{table}

\paragraph{Forecasting accuracy.}
Table~\ref{tab:future-macro} summarizes the nine-state results. Across 20 cutoff--horizon cells per source, \method{} reduces transformed-space MAE from 0.144591 to $0.104015\pm0.000075$ on MIMIC and from 0.106891 to $0.083137\pm0.000063$ on VitalDB, corresponding to 28.06\% and 22.22\% reductions relative to persistence. MAE-Skill increases from 0.0899 to $0.3505\pm0.0005$ and from 0.2669 to $0.4356\pm0.0004$, respectively; $\rho$ increases from 0.4866 to $0.6728\pm0.0004$ and from 0.6333 to $0.7527\pm0.0003$. The nine-state aggregate favors \method{} in MAE, Skill, and $\rho$ in all 40 source--cutoff--horizon cells. This agreement characterizes consistency across the evaluated conditions, not 40 independent replications.

\paragraph{Three-seed forecasting controls.}
The compact ablation in Table~\ref{tab:state-ablation-mini} fixes pretraining seed 0, so we separately test control stability across pretraining seeds $\{0,1,2\}$ at epoch 79. On available-case support, the no-state ridge and capacity-matched MLP controls attain equal-source Skills of $0.2844\pm0.0115$ and $0.3488\pm0.0027$, respectively, while the shuffled-target direct predictor attains $0.0038\pm0.0006$ (Appendix Table~\ref{tab:forecast-controls-three-seed}). Thus, the no-state representation retains recoverable future-state information after fitting a predictor, whereas target shuffling yields a direct predictor that is near zero only after source averaging. Because these controls use available-case support rather than the persistence-paired support of Table~\ref{tab:future-macro}, we do not compare their absolute Skill values directly with the primary \method{} result.

\paragraph{Secondary forecasting diagnostics.}
Additional horizon, cutoff, and ECG-conditional PAT analyses are reported in Appendix~\ref{app:future-eval}. Absolute $H_4$ MAE remains below persistence on both sources, although relative $H_1$--$H_4$ degradation is larger for \method{}; performance is better than persistence at each of the five fixed cutoffs. PAT is treated as a conditional diagnostic because its persistence reference uses an observed ECG-derived quantity and its physical-unit results are source dependent.

\paragraph{Information-access and target-quality checks.}
Across 10,000 audited windows from each source, five stored-suffix interventions produced zero recorded deviation in visible states, registers, forecast context, or predictions; the maximum absolute suffix-input gradient was also zero. These checks support the declared boundary at the stored-window interface, not upstream preprocessing. Appendix~\ref{app:future-eval} gives the full intervention/gradient audit and the separate 20,000-window automated target-quality assessment; extractor coverage is not treated as label correctness.

\section{Discussion and Limitations}

\method{} makes the conditioning set auditable at the stored, preprocessed-window interface. That scope is deliberate: the guarantee does not cover upstream filtering or resampling and therefore does not imply raw-signal streaming equivalence. The intervention and gradient audits test the implemented boundary directly, whereas forecasting accuracy tests a different question---whether the admissible prefix contains useful information about future beat states.

The forecasting controls sharpen the role of state supervision. At seed 0, aligned supervision improves Skill over a capacity-matched no-state predictor, while shuffled targets yield a near-median direct predictor. Across three pretraining seeds, however, ridge and MLP probes fitted to the no-state representation still achieve positive available-case Skill. We therefore interpret future-state supervision as an explicit alignment objective for forecastable physiological descriptors, not as evidence that other pretraining objectives cannot encode future-predictive information. Downstream transfer is likewise task dependent and uses a different checkpoint and readout from intrinsic forecasting.

The deterministic targets remain measurement-dependent signal proxies rather than clinical ground truth. Validity masks and high extractor coverage do not eliminate systematic beat-detection or morphology-extraction error, and the automated ECG-reference audit is not a clinical annotation study. Forecasting evaluation holds out groups within MIMIC and VitalDB rather than entire sources, and the frozen Event-VQ tokenizer predates the MIMIC patient split. Finally, PPG-only inference reduces sensing requirements but not model cost: the backbone remains large for resource-constrained deployment. Extended limitations, forecasting and target-quality audits, and additional transfer/efficiency analyses appear in Appendices~\ref{app:discussion},~\ref{app:future-eval}, and~\ref{app:additional-analyses}.

\section{Conclusion}
\label{sec:conclusion}

We introduced \method{}, a PPG representation learner that couples interpretable future-beat forecasting with an explicit stored-prefix information boundary. By separating what the model may observe from what it is trained to predict, \method{} makes predictive pretraining auditable by construction while retaining PPG-only inference. Across five cutoffs and four future-beat horizons, \method{} reduces nine-state transformed-space MAE relative to persistence by 28.06\% on MIMIC and 22.22\% on VitalDB, with consistent aggregate gains in MAE, Skill, and rank correlation; in a separate seven-model comparison, it attains the best mean on 9/13 frozen linear-probe tasks and 7/13 full-fine-tuning tasks. Boundary interventions and input-gradient audits support the declared stored-window contract, while three-seed controls show that explicit state supervision aligns representations with forecastable physiology rather than uniquely creating predictive information. More broadly, \method{} frames predictive physiological representation learning around three separately testable questions: \emph{what information is accessible, what future structure is supervised, and how the learned representation transfers}.

\clearpage

\subsection*{AI Use Statement}

Generative AI tools were used to organize code-and-log evidence, draft and edit prose and LaTeX, check internal consistency, and provide feedback on the presentation of mathematical arguments, experimental reporting, and interpretation. They were not used to generate experimental measurements, replace model training, invent citations, or determine statistical significance. Mathematical claims, proofs, numerical results, citations, and experimental conclusions were independently checked by the authors, who take responsibility for the final content.

\subsection*{Reproducibility Statement}

Appendices~\ref{app:architecture},~\ref{app:final-eval-protocol},
\ref{app:repro}, and~\ref{app:future-eval} document the model specification,
objective weights, preprocessing interfaces, seed hierarchy, checkpoint
selection, and evaluation/aggregation procedures. Cross-model transfer uses the
validation-selected zero-based epoch-29 EMA checkpoint; future-state evaluation
and stored-suffix audits use the terminal epoch-79 EMA checkpoint from the same
three pretraining runs. Test labels are not used for checkpoint selection,
hyperparameter tuning, or ablation choice. The stored-window information
boundary is evaluated by suffix-intervention and input-gradient audits, while
target-quality checks use an automatically detected ECG reference and are not
clinical ground-truth annotation.

The frozen Event-VQ tokenizer predates the MIMIC patient split, so MIMIC
patient-disjoint claims refer specifically to \method{} backbone pretraining
(Appendix~\ref{app:event-vq-tokenizer}). MIMIC preprocessing is documented by an
executable entry point and corpus fingerprints. For VitalDB, we recovered the
multimodal converter, schema, preprocessing configuration, processing log, and
corpus-level manifest used to construct the reported shards. Appendix~
\ref{app:vitaldb-multimodal} documents track selection, shared-grid windowing,
the stored schema, full-corpus modality/scaffold availability, and an explicit
independent-audit protocol. Raw-record reconstruction exactly reproduced three
sampled stored windows, but this check shares the VitalDB SDK temporal grid and
does not constitute an exhaustive native-timestamp or acquisition-device
synchronization audit.

\bibliography{references}
\bibliographystyle{iclr2027_conference}

\clearpage

\appendix
\section{Extended Discussion and Limitations}
\label{app:discussion}

\paragraph{Information access versus architectural connectivity.}
\method{} separates what a physiological predictor may observe from how it processes those observations. Proposition~1 derives stored-suffix invariance from prefix-only input construction; bidirectional attention over the same inputs would preserve that property. Architectural connectivity therefore governs representation mixing within an already established boundary. The completed stored-suffix intervention and gradient audits produced zero recorded suffix-induced deviations and zero suffix-input gradient under the audited stored-window interface, providing empirical support for this implementation-level contract. This is still a forward-access guarantee, not statistical independence between past and future. Its starting point matters: offline filtering, resampling, and full-window normalization can transmit raw-suffix information into stored prefixes, while prefix-global views do not imply streaming equivalence. A raw-signal guarantee requires boundary-preserving preprocessing and raw-to-context interventions.

\paragraph{Prediction and transfer are distinct outcomes.}
The seed-0 contrasts in Table~\ref{tab:ablation-main} show that forecasting and transfer need not favor the same configuration. Full \method{} improves forecasting Skill over the capacity-matched no-state predictor at that checkpoint, whereas downstream differences are mixed across tasks and interventions. The separate three-seed controls in Table~\ref{tab:forecast-controls-three-seed} further show that no-state representations retain recoverable future-state information when a ridge or MLP predictor is fitted after pretraining. We therefore interpret explicit state supervision as an alignment objective for the evaluated future descriptors rather than as a claim that other objectives cannot encode predictive information. Transfer uses an earlier checkpoint and a different readout from forecasting, so intrinsic Skill is a complementary diagnostic rather than a sufficient explanation of downstream utility.

\paragraph{Inspectable targets remain measurement-dependent.}
Deterministic descriptors expose supervision semantics without making extracted labels ground truth. Validity masking excludes inadmissible targets but does not correct systematic detection errors; coverage is not target accuracy. Horizons index accepted beats rather than fixed durations. Although absolute error remains below persistence through four beats, relative $H_1$--$H_4$ degradation is faster (Table~\ref{tab:future-horizon-degradation}). PAT further illustrates the distinction between proxy and physical metrics: on MIMIC, lower transformed MAE coexists with worse millisecond MAE and rank correlation (Table~\ref{tab:future-pat}). Predictive skill must therefore be interpreted alongside target quality, event timing, and physical-unit error rather than as direct evidence of clinical accuracy.

\paragraph{Generalization and deployment scope.}
The main forecasting tests hold out groups within two pretraining sources, not entire sources; BIDMC also remains overlap-sensitive in transfer comparisons. PPG-only inference concerns sensing requirements, not computational efficiency: the downstream backbone contains 97.0M parameters (Table~\ref{tab:params}) and executes all 12 layers without early termination. Compression and streaming transforms are deployment directions, while device and demographic shifts, calibration, energy use, and prospective clinical validation require separate assessment.

\section{Exact Future-State Targets}
\label{app:targets}

Table~\ref{tab:target-definitions} transcribes the deterministic target implementation. The target extractor first replaces non-finite PPG samples by zero, computes visible-prefix statistics, normalizes and clips to $[-8,8]$, applies a 0.5--8~Hz FFT band-pass, and detects refractory local maxima. The configured cutoff guard is 0.30~s, while onset extraction uses a 0.35-s lookback. At 50~Hz these round to 15 and 18 samples, respectively, so candidate selection uses their maximum: an effective 18-sample (0.36-s) future-peak guard. It also reserves a 0.55-s post-peak window. An onset is the minimum filtered value in the preceding 0.35~s. Invalid elements are zero-filled and excluded from Equation~\ref{eq:state-loss}.

\begin{table}[!htbp]
\caption{Exact state transforms. $\operatorname{clip}(z,a,b)$ truncates
$z$ to $[a,b]$; $o,p$ are onset and peak values; $U,D$ are pulse areas
after and before the peak.}
\label{tab:target-definitions}
\centering
\scriptsize
\setlength{\tabcolsep}{3pt}
\renewcommand{\arraystretch}{1.08}
\begin{tabularx}{\textwidth}{@{}lYY@{}}
\toprule
Code name & Target transform & Element validity \\
\midrule
log\_ibi & $\operatorname{clip}(\log(\operatorname{clip}(\mathrm{IBI},.30,2.0)/.80),-1.25,1)$ & previous peak; IBI $\in[.30,2.0]$ s \\
log\_amplitude & $\operatorname{clip}(\log(1+(p-o)_+),0,2.5)$ & amplitude $>.05$ \\
log\_rise\_time & $\operatorname{clip}(\log(\operatorname{clip}(R,.03,.70)/.20),-1.75,1.25)$ & $R\in(.02,.70)$ s \\
log\_half\_width & $\operatorname{clip}(\log(\operatorname{clip}(W,.03,.90)/.30),-1.75,1.25)$ & $W\in(.02,.90)$ s \\
log\_area & $\operatorname{clip}(\log(1+S),0,2.5)$ & valid beat \\
baseline & $\tanh(o/3)$ & valid beat \\
asymmetry & $\operatorname{clip}((U-D)/(U+D+10^{-5}),-1,1)$ & valid beat \\
haar\_fast & $\operatorname{clip}(\log((E_{d1}+10^{-6})/E_{\rm tot}),-6,2)/3$ & valid beat \\
haar\_mid & $\operatorname{clip}(\log((E_{d2}+10^{-6})/E_{\rm tot}),-6,2)/3$ & valid beat \\
log\_pat & $\operatorname{clip}(\log(\operatorname{clip}(\mathrm{PAT},.05,.65)/.25),-1.75,1.25)$ & paired ECG; PAT $\in[.05,.65]$ s \\
\bottomrule
\end{tabularx}
\end{table}

The ECG trace is centered, standardized, converted to magnitude, and subjected to refractory R-peak detection. For each PPG onset, the most recent R-peak between 0.65 and 0.05~s earlier defines PAT. The data loader supplies a per-window \code{ecg\_available} flag, preventing zero-filled placeholder ECG from becoming a target. PPG-derived target validity is independent of PAT validity.

\section{Architecture and Objective Details}
\label{app:architecture}

\paragraph{Physiological graph bias.}
Each backbone token is a node with a type index $\kappa_i$ and a time index $t_i$.
Registers have $t_i<0$ and one of seven factor types (cardiac, respiratory,
vascular, noise, subject, domain, or task). PPG patches use type
\code{ppg\_raw} with $t_i\in\{0,\ldots,T-1\}$; the six pseudo-variate
streams use types \code{ppg\_raw}, \code{ppg\_d1}, \code{ppg\_d2},
\code{ppg\_bandpass}, \code{envelope}, and \code{spectrum} at the same patch
times. There is no sparse physiological adjacency matrix. Instead, each
attention head has a learned dense type-pair bias
$B^{\mathrm{physio}}\in\mathbb{R}^{39\times39\times12}$ and a learned
T5-style relative-time bias over 32 buckets with maximum distance 256; both are
initialized with truncated $\mathcal{N}(0,0.02^2)$. The optional heartbeat-event
bias is disabled. For attention head $h$,
\begin{equation}
\operatorname{Attn}_h=
\operatorname{softmax}\!\left(
Q_hK_h^\top/\sqrt{d_h}
+B^{\mathrm{physio}}_{\kappa_i,\kappa_j,h}
+\mathbf{1}[t_i\ge0,t_j\ge0]B^{\mathrm{time}}_{\operatorname{bucket}(t_j-t_i),h}
+M_{ij}\right)V_h,
\label{eq:graph-bias-app}
\end{equation}
where $M_{ij}=-\infty$ for disallowed pairs and $0$ otherwise. Register queries
may read every key; a temporal query may read only earlier temporal keys and
itself, excluding register keys and other same-time tokens. Thus the causal
mask, not the learned bias table, defines permitted connections. The type table
is dense; \code{PHYSIO\_ORDER} only determines variate-token ordering. The
no-graph-bias ablation removes these additive bias terms while leaving the
causal mask unchanged.

\begin{table}[!htbp]
\caption{Parameter counts for the reported model configuration.}
\label{tab:params}
\centering
\small
\begin{tabular*}{0.72\textwidth}{@{\extracolsep{\fill}}lr@{}}
\toprule
Component & Parameters \\
\midrule
Downstream-compatible backbone & 97,000,488 \\
Future-state head & 1,488,778 \\
Trainable \method{} model & 98,489,266 \\
Frozen Event-VQ tokenizer & 3,233,864 \\
Complete pretraining object & 101,723,130 \\
\bottomrule
\end{tabular*}
\end{table}

\begin{table}[!htbp]
\caption{Active objective terms and implementation weights. ``Available
only'' means the loss is omitted when its required target is absent.}
\label{tab:losses}
\centering
\small
\setlength{\tabcolsep}{4pt}
\renewcommand{\arraystretch}{1.06}
\begin{tabularx}{0.88\textwidth}{@{}YcY@{}}
\toprule
Term & Weight & Scope \\
\midrule
Discrete PPG token CE & 1.0 & masked patches only \\
Waveform $L_1+0.5\,\mathrm{SmoothL1}_{\beta=1}$ & 0.5 & hidden patches only \\
ECG scaffold & 0.4 & available only \\
Respiration scaffold & 0.3 & available only \\
Blood-pressure scaffold & 0.5 & available only \\
Quality scaffold & 0.3 & available only \\
Future state smooth L1 & 1.0 & valid active elements \\
Subject supervised contrastive & 0.2 & when IDs/pairs exist \\
Ponder regularizer & 1.0 & expected relative exit-depth penalty; no layer skipping \\
Fourier forecast / order & 0 / 0 & disabled \\
\bottomrule
\end{tabularx}
\end{table}

\subsection{Forecast curriculum and exact joint objective}
\label{app:exact-objective}

For normalized training progress $q\in[0,1]$, the four-phase curriculum activates
\begin{equation}
 K(q)=\min\!\left(4,1+\lfloor4q\rfloor\right).
 \label{eq:curriculum}
\end{equation}
Only the supervised horizons change; input sampling, cutoff policy, and target definitions remain fixed. Writing $\widetilde V_{b,h,k}=V_{b,h,k}\mathbf{1}[h\le K(q)]$, the state loss is
\begin{equation}
 \mathcal{L}_{\mathrm{state}}=
 \frac{\sum_{b,h,k}\widetilde V_{b,h,k}\,
 \ell_{\beta}(\hat y_{b,h,k}-y_{b,h,k})}
 {\sum_{b,h,k}\widetilde V_{b,h,k}},\qquad \beta=0.5,
 \label{eq:state-loss}
\end{equation}
where $\ell_{\beta}(e)=e^2/(2\beta)$ for $|e|<\beta$ and $|e|-\beta/2$ otherwise. Valid active elements receive equal weight.

For waveform residual $e_{b,t,p}=\hat x_{b,t,p}-x_{b,t,p}$ and batch size $B$,
\begin{equation}
 \mathcal{L}_{\mathrm{wave}}=\frac{1}{B}\sum_{b=1}^{B}
 \frac{1}{H_bP}\sum_{t=C_b}^{T-1}\sum_{p=0}^{P-1}
 \left(|e_{b,t,p}|+0.5\,\ell_{1}(e_{b,t,p})\right),
 \label{eq:wave-loss}
\end{equation}
where $\ell_1$ is smooth-$L_1$ with $\beta=1$. The complete objective is
\begin{align}
 \mathcal{L}={}&\mathcal{L}_{\mathrm{token}}+0.5\mathcal{L}_{\mathrm{wave}}
 +\mathcal{L}_{\mathrm{state}}+0.4\mathcal{L}_{\mathrm{ECG}}
 +0.3\mathcal{L}_{\mathrm{resp}}+0.5\mathcal{L}_{\mathrm{BP}}\nonumber\\
 &+0.3\mathcal{L}_{\mathrm{quality}}+0.2\mathcal{L}_{\mathrm{subj}}
 +\mathcal{L}_{\mathrm{ponder}}.
 \label{eq:full-objective}
\end{align}
$\mathcal{L}_{\mathrm{token}}$ is frozen Event-VQ morphology-token cross-entropy on hidden patches. Fourier and order-prediction branches are disabled, no teacher network is used, and all 12 layers execute in training and inference.

\paragraph{Ponder regularizer.}
A per-layer halt head $h_\ell$ maps the mean-pooled hidden state after
layer $\ell=1,\ldots,L$ ($L=12$) to a logit. Let
$\lambda_\ell=\sigma(h_\ell)$ for $\ell<L$ and assign all residual mass to the
last layer:
\begin{equation}
p_1=\lambda_1,\qquad
p_\ell=\lambda_\ell\prod_{k=1}^{\ell-1}(1-\lambda_k)\;(\ell<L),\qquad
p_L=1-\sum_{\ell<L}p_\ell.
\label{eq:ponder-prob-app}
\end{equation}
The regularizer is the mean expected relative exit depth,
\begin{equation}
\mathcal{L}_{\mathrm{ponder}}
=\eta\,\mathbb{E}\!\left[\frac{1}{L}\sum_{\ell=1}^{L}\ell p_\ell\right],
\qquad \eta=10^{-2}.
\label{eq:ponder-app}
\end{equation}
It enters Equation~\ref{eq:full-objective} with outer coefficient $1$. The
last halt head is frozen because $p_L$ does not depend on $h_L$. Training and
reported inference always execute all 12 layers; the halt heads never gate the
residual stream and are not used for early exit.

\paragraph{Training-only auxiliary scaffold losses.}
Table~\ref{tab:scaffold-losses} defines the retained auxiliary objectives.
Regression targets use missing value \code{NaN}, classification targets use
$-1$, and invalid targets contribute no loss. During training, ECG, respiration,
and blood-pressure target groups are dropped independently with probability
0.1; quality supervision is never dropped. No ECG waveform-reconstruction loss
is used, and PAT is not part of these scaffold losses.

\begin{table}[!htbp]
\caption{Training-only auxiliary scaffold losses. Readouts are mean-pooled typed registers. $\operatorname{SL1}$ denotes smooth-$L_1$ loss with $\beta=1$. Invalid targets contribute no loss.}
\label{tab:scaffold-losses}
\centering
\scriptsize
\setlength{\tabcolsep}{2.5pt}
\renewcommand{\arraystretch}{1.10}
\begin{tabularx}{\textwidth}{@{}p{1.8cm}Yp{2.2cm}Yp{2.7cm}@{}}
\toprule
Loss (weight) & Target & Readout & Loss function & Availability \\
\midrule
$\mathcal L_{\mathrm{ECG}}$ (0.4)
& Window HR (bpm), RMSSD (ms), and four-class rhythm (sinus/brady/tachy/irregular) from ECG R-peaks; HR/RMSSD fall back to PPG peaks if ECG is absent
& cardiac-register mean
& $\operatorname{SL1}(\widehat{\mathrm{HR}})+\operatorname{SL1}(\widehat{\mathrm{RMSSD}})+\operatorname{CE}(\widehat{\mathrm{rhythm}})$
& ECG for rhythm; HR/RMSSD if ECG or at least three PPG peaks \\
\addlinespace
$\mathcal L_{\mathrm{resp}}$ (0.3)
& Respiratory rate (breaths/min) and eight-bin Hilbert phase at the window end from 0.1--0.5~Hz respiration; rate falls back to PPG modulation
& respiratory-register mean
& $\operatorname{SL1}(\widehat{\mathrm{RR}})+\operatorname{CE}(\widehat{\phi})$
& Respiration for phase; rate if respiration or PPG fallback \\
\addlinespace
$\mathcal L_{\mathrm{BP}}$ (0.5)
& Arterial SBP, DBP, and MAP (mmHg) from ABP peak/trough medians, with $\mathrm{MAP}=\mathrm{DBP}+(\mathrm{SBP}-\mathrm{DBP})/3$
& vascular-register mean, MLP to three scalars
& sum of $\operatorname{SL1}$ over SBP, DBP, and MAP
& ABP present and targets in $[20,300]$~mmHg \\
\addlinespace
$\mathcal L_{\mathrm{quality}}$ (0.3)
& Scalar SQI from PPG spectral energy in 0.5--8~Hz and four-class artifact label (clean/low-perfusion/clipping/motion)
& noise-register mean; sigmoid SQI head
& $\operatorname{SL1}(\widehat{\mathrm{SQI}})+\operatorname{CE}(\hat a)$
& Always; PPG-only and never dropped \\
\addlinespace
$\mathcal L_{\mathrm{subj}}$ (0.2)
& H5 \code{subject\_id}; same-ID windows in a batch are positives
& subject-register mean, $L_2$-normalized
& supervised contrastive, temperature $0.1$; zero if no positive pair
& valid subject ID and at least one same-ID positive in the batch \\
\bottomrule
\end{tabularx}
\end{table}

\subsection{Frozen Event-VQ tokenizer}
\label{app:event-vq-tokenizer}
$\mathcal L_{\mathrm{token}}$ is masked 4096-way cross-entropy on morphology
IDs from a frozen event-aware VQ tokenizer (Event-VQ). Event-VQ is trained in a
separate first stage and then frozen: gradients are disabled, evaluation mode
is used, and EMA codebook updates are locked. Only the morphology stream is a
pretraining target for full \method{}; event, quality, and rhythm streams
supervise tokenizer training only. Event-VQ is this work's tokenizer rather
than a reused SIGMA checkpoint. It extends fixed-patch VQ tokenization with a
six-view convolutional stem, a four-layer local Transformer, and four
normalized EMA codebooks.

The frozen checkpoint is the online tokenizer state from epoch 49 of a
50-epoch run on the full processed MIMIC corpus (SHA-256
\texttt{\seqsplit{0e63bc8da43e00a7033155017f688cc806a11e6fb3d2cc4b3393bf6ba8b87641}}).
This tokenizer stage predates the later patient-level 90/5/5 split and
therefore includes PPG from the 628 MIMIC groups subsequently assigned to
source validation and test partitions. It does not use VitalDB or any
standalone downstream evaluation dataset. The full \method{} backbone is still
trained only on the 90/5/5 training partitions. Consequently, MIMIC
``group-disjoint'' statements in this paper refer to the backbone pretraining
split, not to the frozen tokenizer stage.

\begin{table}[!htbp]
\caption{Provenance of the frozen Event-VQ tokenizer used by $\mathcal L_{\mathrm{token}}$.}
\label{tab:event-vq-provenance}
\centering
\scriptsize
\setlength{\tabcolsep}{3.2pt}
\renewcommand{\arraystretch}{1.08}
\begin{tabularx}{\linewidth}{@{}lY@{}}
\toprule
Item & Frozen setting \\
\midrule
Role in $\mathcal L_{\mathrm{token}}$ & Morphology-ID targets; CE on masked patches; weight 1.0 \\
Architecture & six PPG views $\rightarrow$ convolutional stem $\rightarrow$ Transformer (4 layers, 8 heads, $D=200$) \\
Codebooks (EMA VQ, dimension 64) & morphology 4096; event 512; quality 128; rhythm 512 \\
Parameters & 3,233,864, including frozen codebooks \\
Tokenizer-stage data & MIMIC-III matched; 359 shards; 50~Hz; 30~s windows; patch size 50 \\
Same 90/5/5 train split? & No; trained on the full processed MIMIC corpus \\
VitalDB / standalone downstream data? & No \\
Checkpoint & epoch 49/50; online tokenizer weights \\
Freezing during \method{} pretraining & gradients off; evaluation mode; codebook EMA locked \\
\bottomrule
\end{tabularx}
\end{table}

The subject contrastive objective can be zero when a batch contains no repeated subject. The loader does not implement subject-balanced sampling. This is a limitation of the retained objective, not evidence that the corresponding register necessarily encodes identity.

\section{Evaluation Protocol and Downstream Analyses}
\label{app:final-eval-protocol}

This section records the canonical configuration of the three full \method{} pretraining runs. Ablations retain this configuration except for the intervention explicitly stated for each condition.
\subsection{Canonical full-model protocol}

Table~\ref{tab:final-protocol} defines the three recorded full-model pretraining runs. Test labels are never used for checkpoint selection, hyperparameter tuning, early stopping, or ablation choice.
\begin{table}[!htbp]
\caption{Canonical training protocol for the three full \method{} runs.}
\label{tab:final-protocol}
\centering
\scriptsize
\setlength{\tabcolsep}{3.2pt}
\renewcommand{\arraystretch}{1.06}
\begin{tabularx}{\textwidth}{@{}lYlY@{}}
\toprule
Category & Setting & Category & Setting \\
\midrule
Backbone & width 768; 12 layers; 12 heads; drop-path 0.1 & Attention & causal; 20 typed registers; graph bias on \\
Input & 30 s at 50 Hz; 1500 samples & Patch size & 50 samples (1 s) \\
Pretraining sources & MIMIC-III matched + VitalDB & Source mixture & 80:20 each epoch \\
Patient split & group-disjoint 90/5/5 & Pretraining seeds & $\{0,1,2\}$ \\
Windows / epoch & 5,160,960 & Source counts / epoch & 4,128,768 MIMIC + 1,032,192 VitalDB \\
Mask & content-independent suffix; ratio 0.30--0.45 & Context / suffix minima & 12 / 8 patches \\
Waveform objective & hidden-only $L_1+0.5\,\mathrm{SmoothL1}_{\beta=1}$; weight 0.5 & Normalization & visible-prefix z-score; scale floor 0.05 \\
State objective & four future beats; weight 1.0 & Guard / curriculum & configured 0.30 s; effective 18 samples (0.36 s); four equal-duration phases progressively activating $H_1$--$H_4$ \\
Optimizer & fused AdamW; $\beta=(0.9,0.95)$; WD 0.05 & LR schedule & peak $6\times10^{-4}$; floor $10^{-5}$; 10-epoch linear warmup from 0, then cosine \\
Training & 80 epochs; grad clip 1.0; BF16 & Hardware & 2 nodes $\times$ 8 A100 \\
Per-GPU batch & 128 & Gradient accumulation & 2 \\
Global batch & 4096 & EMA & 0.9998; tokenizer excluded \\
Downstream split seed & 42 & Run length & 201,600 forward/backward micro-steps (100,800 optimizer updates) \\
Disabled branches & Fourier and order prediction; no teacher network & Boundary scope & stored preprocessed window onward \\
\bottomrule
\end{tabularx}
\end{table}

\begin{table}[!htbp]
\caption{Patient-level pretraining split inventory for the final runs.}
\label{tab:final-pretraining-splits}
\centering
\small
\setlength{\tabcolsep}{5pt}
\begin{tabular}{@{}llrr@{}}
\toprule
Source & Split & Patients & Windows \\
\midrule
\multirow{3}{*}{MIMIC-III matched}
& Train & 5,654 & 16,090,545 \\
& Validation & 314 & 843,424 \\
& Test & 314 & 1,000,765 \\
\midrule
\multirow{3}{*}{VitalDB}
& Train & 5,292 & 1,032,196 \\
& Validation & 294 & 58,962 \\
& Test & 294 & 58,189 \\
\bottomrule
\end{tabular}
\end{table}

All complete-system primary analyses use the same three 80-epoch \method{} pretraining runs. Downstream LP/FT uses split seed 42 and downstream seeds $\{42,43,44\}$ with the zero-based epoch-29 EMA checkpoint selected by a validation-only LP composite on eight development tasks. Future-state evaluation and the stored-suffix boundary audit use the terminal zero-based epoch-79 EMA checkpoint from those same runs, without test-based selection. The MIMIC-only source control matches total sampled windows, optimizer updates, checkpoint index, model capacity, learning-rate schedule, and seed set. The cutoff is sampled independently of waveform, ECG, metadata, detected beats, and target values.

\subsection{Downstream tasks and splits}

Table~\ref{tab:downstream-protocol-app} defines the 13 downstream tasks, primary metrics, and outer evaluation splits. All compared methods use identical evaluation examples, labels, splits, and task metrics. Baseline-specific preprocessing is retained only when required for checkpoint compatibility.

\begin{table}[!htbp]
\caption{Final downstream evaluation protocol. All methods share the same examples, labels, outer splits, and primary metric. LP is primary and FT is secondary.}
\label{tab:downstream-protocol-app}
\centering
\small
\setlength{\tabcolsep}{4.5pt}
\begin{tabular}{lllc}
\toprule
Dataset & Task & Primary metric & Outer evaluation \\
\midrule
\multirow{3}{*}{BIDMC} & Heart rate & MAE $\downarrow$ & subject-level folds \\
& Respiratory rate & MAE $\downarrow$ & subject-level folds \\
& SpO$_2$ & MAE $\downarrow$ & subject-level folds \\
\midrule
\multirow{4}{*}{PPG-BP} & Systolic BP & MAE $\downarrow$ & subject-level folds \\
& Diastolic BP & MAE $\downarrow$ & subject-level folds \\
& Average heart rate & MAE $\downarrow$ & subject-level folds \\
& Hypertension & AUROC $\uparrow$ & subject-level folds \\
\midrule
\multirow{3}{*}{WESAD} & Stress & AUROC $\uparrow$ & LOSO \\
& Affect & F1 $\uparrow$ & LOSO \\
& Emotion & AUROC $\uparrow$ & LOSO \\
\midrule
DaLiA & Activity & F1 $\uparrow$ & LOSO \\
Real-World PPG & Subject identification & F1 $\uparrow$ & fixed sample-stratified split \\
Stanford & Signal quality & macro-AUROC $\uparrow$ & documented approximation \\
\bottomrule
\end{tabular}
\end{table}

Real-World PPG is interpreted only as closed-set identification because subject identity is the class label; a subject-held-out split would change the class space. Baseline-specific frontends may retain the sampling rate, normalization, and input length required by the released checkpoint, but may not alter the evaluated sample set, labels, outer split, or primary metric. The primary clean cross-model aggregate excludes BIDMC HR/RR/SpO$_2$ when a baseline cannot rule out pretraining overlap; all task-level values remain reported.

\subsection{Complete forecasting and downstream ablation analysis}
\label{app:complete-ablation}

\paragraph{Ablation protocol.}
Table~\ref{tab:ablation-main} distinguishes two evaluation scopes.
Downstream linear probing (LP) uses pretraining seeds
$\{0,1,2\}$ and zero-based checkpoint epoch 29.
For each pretraining seed, we first average the results over downstream
seeds $\{42,43,44\}$; we then report the mean and sample standard
deviation across the three pretraining-seed averages.
The table presents seven conditions and five representative LP tasks.
Its forecasting column remains restricted to pretraining seed 0 at
zero-based epoch 79 and therefore does not estimate variation across
pretraining seeds. A separate three-seed forecasting-control analysis is
reported below on available-case support; it is not numerically merged with
the paired-support seed-0 column.

The no-state control removes state supervision; its forecasting
evaluation fits a capacity-matched MLP to the frozen representation.
The shuffled-target control shuffles future-state targets across
examples and evaluates its trained forecasting head directly.
The no-PAT variant removes PAT supervision, the no-curriculum variant
activates all four horizons from the first update, the matched raw-view
control replaces physiological views with token- and parameter-matched
raw views, and the MIMIC-only control removes VitalDB under the
matched-compute protocol.
These interventions evaluate forecasting and transfer separately:
the two evaluations use different checkpoint epochs, and the no-state
forecasting control uses a separately fitted predictor.

\begin{table}[!htbp]
\caption{Ablations with seed-0 forecasting and three-pretraining-seed
downstream evaluation.
Forecast Skill uses patient-macro MAE relative to the training-median
reference, followed by equal weighting over nine PPG-derived states,
five cutoffs, four horizons, and two sources.
LP entries report mean $\pm$ sample standard deviation across
pretraining seeds $\{0,1,2\}$, after averaging downstream seeds
$\{42,43,44\}$ within each pretraining seed.
LP uses epoch 29 and forecasting uses epoch 79, both zero-based.
The no-state forecasting result uses a capacity-matched MLP\@.
Bold identifies the best unrounded mean among the displayed conditions,
not statistical significance; rounded values can therefore appear tied.}
\label{tab:ablation-main}
\centering
\scriptsize
\setlength{\tabcolsep}{2.5pt}
\renewcommand{\arraystretch}{1.12}
\resizebox{\linewidth}{!}{%
\begin{tabular}{@{}lrrrrrr@{}}
\toprule
Condition
& \makecell{Future-state\\macro Skill$\uparrow$\\(seed 0)}
& \makecell{BIDMC HR\\MAE$\downarrow$}
& \makecell{PPG-BP SBP\\MAE$\downarrow$}
& \makecell{WESAD Stress\\AUROC$\uparrow$}
& \makecell{DaLiA Activity\\macro-F1$\uparrow$}
& \makecell{Stanford Quality\\macro-AUROC$\uparrow$} \\
\midrule
Full \method{}
& 0.3926
& $\mathbf{1.974}\pm0.559$
& $14.913\pm0.295$
& $\mathbf{0.953}\pm0.007$
& $0.290\pm0.008$
& $\mathbf{0.934}\pm0.005$ \\
No-state control
& 0.3527
& $2.259\pm0.430$
& $15.022\pm0.498$
& $0.953\pm0.003$
& $0.298\pm0.022$
& $0.931\pm0.007$ \\
Shuffled-target control
& 0.0032
& $2.866\pm0.341$
& $14.736\pm0.156$
& $0.947\pm0.015$
& $0.293\pm0.010$
& $0.925\pm0.004$ \\
No-PAT variant
& \textbf{0.3940}
& $2.899\pm0.183$
& $14.652\pm0.163$
& $0.942\pm0.006$
& $0.305\pm0.009$
& $0.932\pm0.005$ \\
No-curriculum variant
& 0.3909
& $2.214\pm0.419$
& $15.122\pm0.680$
& $0.953\pm0.003$
& $\mathbf{0.313}\pm0.008$
& $0.932\pm0.007$ \\
Matched raw-view control
& 0.3911
& $3.053\pm0.115$
& $15.025\pm0.274$
& $0.948\pm0.003$
& $0.313\pm0.011$
& $0.927\pm0.006$ \\
MIMIC-only control
& 0.3711
& $2.401\pm0.355$
& $\mathbf{14.630}\pm0.154$
& $0.937\pm0.016$
& $0.300\pm0.006$
& $0.933\pm0.002$ \\
\bottomrule
\end{tabular}%
}
\vspace{2pt}

\parbox{\linewidth}{\scriptsize
\textit{Scope.}
LP standard deviations describe variation across three
pretraining-seed averages, not nine independent pretraining runs
or confidence intervals.
The forecasting column remains a single-pretraining-seed comparison.
BIDMC is retained as an overlap-sensitive task, not clean-transfer
evidence. No clean-set aggregate ranking is inferred from this
five-task subset.
}
\end{table}

\paragraph{State alignment and transfer.}
At pretraining seed 0, full \method{} achieves macro Skill 0.3926,
compared with 0.3527 for the capacity-matched no-state MLP control
and 0.0032 for the shuffled-target control.
The full \method{} versus no-state difference is 0.0399 Skill units.
However, capacity matching does not make this a shared-probe
comparison of representation quality, and the near-median result
of the shuffled-target predictor does not establish that its
backbone contains no future-state information.

The three-pretraining-seed LP results provide a broader assessment
of transfer and qualify the earlier fixed-seed pattern.
Compared with the no-state control, full \method{} has lower mean
BIDMC HR MAE (1.974 versus 2.259), lower SBP MAE
(14.913 versus 15.022), and higher quality macro-AUROC
(0.934 versus 0.931), but lower activity macro-F1
(0.290 versus 0.298).
Stress AUROC is nearly identical: the difference in unrounded
means is approximately 0.00034, and both round to 0.953.
Compared with shuffled-target pretraining, full \method{} has better
means on HR, stress, and quality, whereas the shuffled-target
control has lower SBP MAE and slightly higher activity macro-F1.
Thus, the forecasting advantage of aligned state supervision
does not translate into uniformly better downstream performance.
These descriptive differences are not claims of statistical
significance.

\paragraph{Auxiliary targets, views, and sources.}
Removing PAT yields seed-0 Skill 0.3940 versus 0.3926 for the full
model, providing no positive aggregate forecasting evidence for PAT
supervision in this comparison.
Other ECG supervision remains active, so this intervention is not
a PPG-only-pretraining control.
Its downstream effects are mixed: removing PAT improves mean SBP
MAE and activity macro-F1, but worsens HR MAE, stress AUROC, and
quality macro-AUROC.

Removing the curriculum or replacing physiological views produces
numerically close seed-0 Skills of 0.3909 and 0.3911, respectively.
Across three pretraining seeds, both alternatives improve activity
macro-F1 relative to the full model.
The no-curriculum variant has the highest unrounded activity mean,
although both alternatives round to 0.313.
Full \method{} has better mean HR, SBP, and quality results than
either alternative; its stress advantage over the no-curriculum
variant is negligible at the reported precision.
These results support task-dependent effects rather than a
universal benefit from the curriculum or physiological views.

Under the matched-compute source comparison, full \method{} exceeds
the MIMIC-only control by 0.0215 seed-0 aggregate Skill.
Its three-pretraining-seed mean stress AUROC is higher by
approximately 0.0162, and its HR MAE is lower
(1.974 versus 2.401).
Conversely, MIMIC-only pretraining yields lower SBP MAE
(14.630 versus 14.913) and higher activity macro-F1
(0.300 versus 0.290); the quality difference is small
(0.933 versus 0.934).
The source-averaged forecasting contrast does not identify which
evaluation source accounts for the gain.

Overall, the multi-seed transfer evaluation strengthens
the characterization of variability while retaining a mixed
component-level picture.
The results distinguish conditional descriptor forecasting from
downstream transfer, rather than demonstrating uniform gains from
every design choice.
They do not isolate future-state versus visible-state supervision
or establish the necessity of the attention pattern.

\paragraph{Three-seed forecasting controls.}
We additionally evaluate three forecasting controls using pretraining seeds
$\{0,1,2\}$ and zero-based checkpoint epoch 79: a ridge predictor and a
capacity-matched MLP fitted to the frozen no-state representation, and the
forecasting head trained with shuffled future-state targets. For each seed,
we average available-case nine-state patient-macro MAE-Skill equally over
five cutoffs and four horizons. The equal-source aggregate gives MIMIC and
VitalDB equal weight within each seed; entries report means and sample
standard deviations across the three pretraining seeds. These controls use a
different support definition from the persistence-paired main forecasting
comparison and are therefore reported separately.

\begin{table}[!htbp]
\caption{Three-seed forecasting controls at zero-based epoch 79.
Values are available-case nine-state macro MAE-Skill relative to the
training-median reference. Within each seed, scores average five cutoffs and
four horizons; the aggregate weights MIMIC and VitalDB equally. Entries report
mean $\pm$ sample SD across pretraining seeds $\{0,1,2\}$. This support differs
from the persistence-paired main forecasting comparison.}
\label{tab:forecast-controls-three-seed}
\centering
\scriptsize
\setlength{\tabcolsep}{4pt}
\begin{tabular*}{\linewidth}{@{\extracolsep{\fill}}lrrr@{}}
\toprule
Control & MIMIC & VitalDB & Equal-source aggregate \\
\midrule
No-state + ridge
& $0.2592\pm0.0091$ & $0.3095\pm0.0140$ & $0.2844\pm0.0115$ \\
No-state + capacity-matched MLP
& $0.3073\pm0.0049$ & $0.3904\pm0.0012$ & $0.3488\pm0.0027$ \\
Shuffled-target direct predictor
& $0.0709\pm0.0004$ & $-0.0632\pm0.0011$ & $0.0038\pm0.0006$ \\
\bottomrule
\end{tabular*}
\end{table}

The positive Skill of both no-state probes shows that future-state information
is recoverable from a representation learned without the explicit state
objective. The capacity-matched MLP outperforms the ridge probe, indicating
that predictor capacity matters for this diagnostic. The shuffled-target
direct predictor is near zero only after source averaging: it remains positive
on MIMIC and negative on VitalDB. These results characterize the evaluated
predictors; they neither establish an information limit of the underlying
representations nor replace an independently trained causal forecasting
baseline.

\FloatBarrier

\subsection{Statistical aggregation}

Aggregation follows the source of stochastic variation in each experiment family. For complete-system comparisons, downstream seeds are first averaged within each \method{} pretraining seed, after which we report the mean and sample standard deviation across pretraining seeds. Released baselines provide variation over their available downstream runs. Seed-0 ablations and label-efficiency experiments instead fix the pretrained checkpoint and summarize downstream seeds only; these repetitions are therefore not treated as independent pretraining replications.

With three independent pretraining seeds, we use variability estimates descriptively and do not interpret sample standard deviations as confidence intervals. Folds, windows, crops, and test-time views are likewise not treated as independent training replicates.

\subsection{Cross-model downstream aggregate}

The cross-model aggregate is complete for seven models. \method{} uses
the validation-only selected epoch-29 EMA checkpoint. For \method{},
each pretraining seed first averages downstream seeds 42/43/44; the
reported sample SD is then computed across pretraining seeds. Released
baselines use the three downstream seeds directly.

Across the 26 non-random-initialization LP/FT task--protocol cells, \method{} attains the best mean in 16; across the 20 non-BIDMC cells its mean rank is 1.45. These counts summarize task-level outcomes and are not treated as independent replications.

\subsubsection*{Full fine-tuning results}
The primary paper reports frozen linear probing; Table~\ref{tab:cross-model-ft} gives the secondary full fine-tuning comparison on the same 13 tasks.

\begin{table}[!htbp]
\caption{Cross-model downstream performance across the same 13 PPG tasks under full fine-tuning. The average columns are descriptive arithmetic means of task means and are shown without cross-task uncertainty.}
\label{tab:cross-model-ft}
\centering
\scriptsize
\setlength{\tabcolsep}{1.2pt}
\renewcommand{\arraystretch}{1.12}

\begin{tabular*}{\linewidth}{@{\extracolsep{\fill}}lccccccc@{}}
\multicolumn{8}{c}{\textit{Regression tasks (MAE $\downarrow$)}}\\[-1.5pt]
\toprule
& \multicolumn{3}{c}{BIDMC} & \multicolumn{3}{c}{PPG-BP} & \\
\cmidrule(lr){2-4}\cmidrule(lr){5-7}\cmidrule(lr){8-8}
Method & HR & RR & SpO$_2$ & SBP & DBP & HR & Avg. \\
\midrule
PaPaGei-S
& 1.796$\pm$0.137 & 2.391$\pm$0.072 & 3.138$\pm$0.108
& 16.752$\pm$0.149 & 8.903$\pm$0.026 & 8.804$\pm$0.075 & 6.964 \\
PaPaGei-P
& 2.080$\pm$0.039 & 2.331$\pm$0.071 & 2.883$\pm$0.061
& 16.412$\pm$0.173 & 8.906$\pm$0.123 & 8.432$\pm$0.088 & 6.841 \\
PaPaGei-S-SVRI
& 2.227$\pm$0.188 & 2.326$\pm$0.073 & 2.883$\pm$0.180
& 16.841$\pm$0.220 & 8.964$\pm$0.073 & 8.759$\pm$0.081 & 7.000 \\
AnyPPG
& 1.932$\pm$0.204 & 2.274$\pm$0.116 & 2.734$\pm$0.128
& \textbf{15.362$\pm$0.770} & 8.977$\pm$0.265 & 5.308$\pm$0.347 & 6.098 \\
Pulse-PPG
& 2.270$\pm$0.121 & 2.956$\pm$0.109 & 3.339$\pm$0.219
& 17.612$\pm$0.435 & 9.628$\pm$0.480 & 7.997$\pm$0.925 & 7.300 \\
SIGMA-PPG
& \textbf{1.758$\pm$0.116} & 2.510$\pm$0.041 & 3.013$\pm$0.239
& 15.984$\pm$0.414 & 8.882$\pm$0.036 & 4.531$\pm$0.187 & 6.113 \\
\midrule
\textbf{\method{}}
& 2.025$\pm$0.093 & \textbf{2.273$\pm$0.028}
& \textbf{2.500$\pm$0.069} & 15.677$\pm$0.414
& \textbf{8.797$\pm$0.050} & \textbf{4.376$\pm$0.129} & \textbf{5.941} \\
\bottomrule
\end{tabular*}

\vspace{5pt}

\begin{tabular*}{\linewidth}{@{\extracolsep{\fill}}lcccccccc@{}}
\multicolumn{9}{c}{\textit{Classification and recognition tasks ($\uparrow$)}}\\[-1.5pt]
\toprule
& PPG-BP & \multicolumn{3}{c}{WESAD} & DaLiA & Real-World & Stanford & \\
\cmidrule(lr){2-2}\cmidrule(lr){3-5}\cmidrule(lr){6-6}
\cmidrule(lr){7-7}\cmidrule(lr){8-8}\cmidrule(lr){9-9}
Method
& \makecell{HTN\\AUC}
& \makecell{Stress\\AUC}
& \makecell{Affect\\F1}
& \makecell{Emotion\\AUC}
& \makecell{Act.\\F1}
& \makecell{ID\\F1}
& \makecell{Qual.\\mAUC} & Avg. \\
\midrule
PaPaGei-S
& 0.492$\pm$0.014 & 0.947$\pm$0.002 & 0.529$\pm$0.010
& 0.519$\pm$0.009 & 0.293$\pm$0.004 & 0.862$\pm$0.014
& 0.874$\pm$0.007 & 0.645 \\
PaPaGei-P
& 0.525$\pm$0.028 & \textbf{0.966$\pm$0.001} & 0.552$\pm$0.012
& 0.546$\pm$0.011 & 0.312$\pm$0.003 & 0.868$\pm$0.018
& 0.893$\pm$0.003 & 0.666 \\
PaPaGei-S-SVRI
& 0.497$\pm$0.035 & 0.952$\pm$0.003 & 0.554$\pm$0.003
& 0.571$\pm$0.031 & 0.312$\pm$0.008 & 0.796$\pm$0.026
& 0.892$\pm$0.003 & 0.653 \\
AnyPPG
& \textbf{0.693$\pm$0.007} & 0.965$\pm$0.001 & 0.523$\pm$0.011
& 0.551$\pm$0.024 & 0.315$\pm$0.004 & 0.875$\pm$0.014
& 0.934$\pm$0.001 & 0.694 \\
Pulse-PPG
& 0.537$\pm$0.021 & 0.943$\pm$0.005 & 0.544$\pm$0.007
& 0.586$\pm$0.027 & \textbf{0.351$\pm$0.001}
& \textbf{0.959$\pm$0.011} & 0.951$\pm$0.003 & 0.696 \\
SIGMA-PPG
& 0.535$\pm$0.026 & 0.949$\pm$0.005 & 0.565$\pm$0.013
& 0.536$\pm$0.025 & 0.325$\pm$0.005 & 0.810$\pm$0.025
& 0.947$\pm$0.005 & 0.667 \\
\midrule
\textbf{\method{}}
& 0.610$\pm$0.021 & 0.960$\pm$0.003
& \textbf{0.593$\pm$0.007} & \textbf{0.592$\pm$0.017}
& 0.344$\pm$0.002 & 0.888$\pm$0.009
& \textbf{0.954$\pm$0.001} & \textbf{0.706} \\
\bottomrule
\end{tabular*}

\vspace{2pt}

\end{table}

\begin{table}[!htbp]
\caption{SIGMA-PPG-style descriptive averages for the seven-model
downstream comparison. Regression averages are raw arithmetic means of
the six MAE task means ($\downarrow$); classification/recognition averages
are arithmetic means of the seven task-specific bounded metrics
($\uparrow$). Because these averages combine heterogeneous tasks (and, for
regression, different physical units), they are descriptive summaries rather
than scale-normalized effect estimates; no cross-task uncertainty is attached.
Bold marks the best average in each column.}
\label{tab:formal-task-averages-app}
\centering
\scriptsize
\setlength{\tabcolsep}{3.0pt}
\renewcommand{\arraystretch}{1.10}
\begin{tabular*}{\linewidth}{@{\extracolsep{\fill}}lcccc@{}}
\toprule
Method
& \makecell{LP Regression\\Avg.$\downarrow$}
& \makecell{LP Class./Recog.\\Avg.$\uparrow$}
& \makecell{FT Regression\\Avg.$\downarrow$}
& \makecell{FT Class./Recog.\\Avg.$\uparrow$} \\
\midrule
PaPaGei-S       & 7.356 & 0.452 & 6.964 & 0.645 \\
PaPaGei-P       & 7.518 & 0.542 & 6.841 & 0.666 \\
PaPaGei-S-SVRI  & 7.583 & 0.470 & 7.000 & 0.653 \\
AnyPPG          & 7.324 & 0.590 & 6.098 & 0.694 \\
Pulse-PPG       & 8.745 & 0.587 & 7.300 & 0.696 \\
SIGMA-PPG       & 6.816 & 0.518 & 6.113 & 0.667 \\
\midrule
\textbf{\method{}} &
\textbf{5.919} &
\textbf{0.650} &
\textbf{5.941} &
\textbf{0.706} \\
\bottomrule
\end{tabular*}
\end{table}

\begin{table}[!htbp]
\caption{Validation-only checkpoint selection over 16 EMA candidates.
Composite is the prespecified eight-task LP rank composite. Bold marks
the selected checkpoint.}
\label{tab:checkpoint-selection-app}
\centering
\small
\setlength{\tabcolsep}{8pt}
\renewcommand{\arraystretch}{1.06}
\begin{tabular}{rr@{\qquad}rr}
\toprule
Epoch & Composite & Epoch & Composite \\
\midrule
4  & 0.2167 & 44 & 0.3583 \\
9  & 0.5833 & 49 & 0.3833 \\
14 & 0.5500 & 54 & 0.4083 \\
19 & 0.5167 & 59 & 0.3667 \\
24 & 0.6167 & 64 & 0.4333 \\
\textbf{29} & \textbf{0.7333} & 69 & 0.5333 \\
34 & 0.6500 & 74 & 0.5750 \\
39 & 0.5250 & 79 & 0.5500 \\
\bottomrule
\end{tabular}
\end{table}

\subsection{Selected quantitative summary}

\paragraph{Reporting convention.}
A dash denotes a quantity for which no pooled cross-source aggregate is reported. Forecasting statistics use support shared by \method{} and persistence;
they therefore need not equal available-case statistics reported elsewhere.
Downstream comparisons and ablation results are reported separately.

\begin{table}[!htbp]
\caption{Selected quantitative summary.
Forecasting values are means over three pretraining seeds; reported
standard deviations are sample standard deviations across seed-level
aggregates.
MIMIC and VitalDB receive equal weight in the forecasting aggregate.
Automatic target audits measure agreement with an ECG-based reference
and do not provide clinical ground-truth annotation.
Efficiency entries report architecture-only encoder inference measurements under the common benchmark in Appendix~\ref{app:additional-analyses}.}
\label{tab:final-results-sheet-app}
\centering
\scriptsize
\setlength{\tabcolsep}{3pt}
\renewcommand{\arraystretch}{1.06}

\begin{tabularx}{\textwidth}{
@{}p{2.35cm}>{\raggedright\arraybackslash}Xrrr@{}}
\toprule
Evidence & Quantity & MIMIC & VitalDB & Aggregate / final \\
\midrule

Stored-window boundary
& max $\|\Delta\mathrm{context}\|_{\infty}$
& 0.0 & 0.0 & 0.0 \\
& max $\|\Delta\mathrm{prediction}\|_{\infty}$
& 0.0 & 0.0 & 0.0 \\
& max suffix-input gradient
& 0.0 & 0.0 & 0.0 \\

\midrule
Intrinsic forecasting
& nine-state macro MAE-Skill
& $0.3505\pm0.0005$
& $0.4356\pm0.0004$
& $0.3930\pm0.0004$ \\
& conditional PAT MAE-Skill
& $0.0545\pm0.0002$
& $0.1171\pm0.0005$
& $0.0858\pm0.0003$ \\
& horizon-4 minus horizon-1 Skill
& $-0.0622$ & $-0.0487$ & $-0.0554$ \\

Forecast support
& paired PPG-state coverage (\%)
& 99.66 & 99.18 & 99.42 \\
& paired PAT coverage (\%)
& 94.85 & 89.03 & 91.94 \\

\midrule
Automatic target-quality assessment
& sampled windows
& 10,000 & 10,000 & 20,000 \\
& ECG-available windows
& 3,785 & 10,000 & 13,785 \\
& Matched PPG--ECG interval MAE (ms)
& 82.65 & 73.48 & -- \\
& ECG-reference beat sensitivity (\%)
& 55.52 & 50.71 & -- \\
& ECG-reference beat PPV (\%)
& 73.50 & 81.27 & -- \\

\midrule
Downstream transfer
& task--protocol cells with best mean
& & & 16/26 \\
& LP regression descriptive mean
& & & 5.919 \\
& LP class./recognition descriptive mean
& & & 0.650 \\
& FT regression descriptive mean
& & & 5.941 \\
& FT class./recognition descriptive mean
& & & 0.706 \\
& wins/losses vs.\ SIGMA-PPG
& & & 24/2 \\
& non-BIDMC LP/FT mean rank
& & & 1.45 \\

\midrule
Efficiency
& Full \method{} encoder inference throughput, B64 (windows/s)
& & & 1549.59 \\
& Full \method{} peak allocated inference memory, B64 (GB)
& & & 1.286 \\
& Full \method{} mean batch-1 encoder latency (ms)
& & & 14.370 \\
& Full \method{} mean batch-64 encoder latency (ms)
& & & 41.301 \\

\bottomrule
\end{tabularx}

\vspace{3pt}
\parbox{\linewidth}{\scriptsize
\textit{Forecasting.}
Skill and coverage use the intersection of valid targets, persistence,
and model predictions.
Source-specific Skill averages nine PPG states, five cutoffs, and
four horizons; PAT is reported separately.
PAT coverage is conditional on ECG availability.
The horizon difference uses paired support separately at each horizon,
rather than a fixed intersection across all horizons.
Coverage indicates availability, not target correctness.

\smallskip
\textit{Transfer and ablations.}
Task-level ranks and descriptive means summarize the seven-model comparison. The non-BIDMC mean rank averages 20 task--protocol cells across seven models.
Regression descriptive means combine different units and are not
scale-normalized measures of overall performance.
Classification/recognition means likewise combine distinct metrics.
Mean task-level standard deviations are not reported as uncertainty
for these aggregates.
Pooled paired effects remain unreported because no common
cross-task effect scale has been specified.

\smallskip
\textit{Target quality and efficiency.}
The automatic audit is complete and uses an ECG-derived reference rather than clinical ground truth.
Inference measurements use an A100-SXM4-80GB, BF16 eager execution,
100 warm-up iterations, and 1,000 measured iterations, without loading
pretrained weights.
Memory denotes peak allocated inference memory in decimal GB,
not training memory.
The efficiency rows above are architecture-only encoder measurements and are not used as training-cost estimates.
}
\end{table}

\section{Data and Training Reproducibility}
\label{app:repro}

\subsection{MIMIC preprocessing and corpus fingerprints}
\label{app:mimic-preprocessing}

MIMIC preprocessing is defined by an executable entry point
\footnote{\code{scripts/preprocess\_mimic\_full.sh}, invoking
\code{preprocessing\_main.py --dataset\_name mimic\_iii\_mm}.}
The pipeline resamples to 50~Hz, forms 30-s windows and 1-s patches, and stores
aligned ECG, respiration, and arterial-pressure-derived metadata when
available. Site-specific data roots are intentionally omitted. The loader
discovers shards non-recursively and fails closed on unsupported recursive
filename patterns.

\begin{table}[!htbp]
\caption{Pretraining data inventory and SHA-256 fingerprints recorded by the
\method{} manifest path.}
\label{tab:data-fingerprints}
\centering
\scriptsize
\setlength{\tabcolsep}{3pt}
\renewcommand{\arraystretch}{1.06}
\begin{tabularx}{\textwidth}{@{}lrrY@{}}
\toprule
Source & Shards & Windows & Fingerprint \\
\midrule
MIMIC & 359 & 17,934,734 &
\texttt{\seqsplit{99e907e4270c90852c54343c392a752dc96f63580e44a8fb165b842adade3c91}} \\
VitalDB & 114 & 1,149,347 &
\texttt{\seqsplit{837adc8937e0282f303ac83da4e3bc9b297fa459e5f4e6be733dbe4ff045bc22}} \\
\bottomrule
\end{tabularx}
\end{table}

The final runs draw 5,160,960 windows per epoch: 4,128,768 from MIMIC and
1,032,192 from VitalDB, preserving the 80:20 source mixture within the fixed
patient-level training partitions in Table~\ref{tab:final-pretraining-splits}.

\FloatBarrier

\subsection{VitalDB multimodal conversion, provenance, and audit}
\label{app:vitaldb-multimodal}

\paragraph{Recovered conversion path and corpus provenance.}
The multimodal VitalDB corpus used for the reported runs was produced by a
dedicated historical conversion path, not by the PPG-only
\code{VitaldbProcessor} in the initial source bundle. We recovered the
converter, HDF5 schema, preprocessing entry point, processing log, and
corpus-level conversion manifest. The recovered manifest records 6,388 discovered case records,
of which 6,153 completed conversion and 235 failed, producing 114 HDF5
shards and 1,149,347 stored 30-s windows. These case counts refer to
conversion records and are not interpreted as numbers of independent
patients without an additional subject-mapping audit.

Table~\ref{tab:vitaldb-conversion-contract} summarizes the recorded
conversion contract. The source grid is sampled at 125~Hz, while stored
waveforms contain 1,500 samples per 30-s window at 50~Hz. The stored
\code{window\_start\_sample} field is an index on the 125-Hz source grid,
not on the 50-Hz stored waveform. The recovered converter has SHA-256
\texttt{\seqsplit{e9a00df30edccfe561aaf2a81edd55eaa2c301d525b0498de617dce172ad58f1}}.
We report this hash for provenance of the recovered implementation; it
does not, by itself, establish that every historical shard was generated
from a byte-identical source snapshot.

\begin{table}[!htbp]
\caption{Recorded contract of the multimodal VitalDB conversion used to
construct the reported corpus. Case counts describe conversion records,
not independently verified patient identities.}
\label{tab:vitaldb-conversion-contract}
\centering
\small
\setlength{\tabcolsep}{5pt}
\begin{tabular}{@{}lr@{}}
\toprule
Item & Recorded value \\
\midrule
Discovered case records & 6,388 \\
Completed case records & 6,153 \\
Failed case records & 235 \\
Output shards & 114 \\
Stored windows & 1,149,347 \\
Source-grid sampling rate & 125 Hz \\
Stored-waveform sampling rate & 50 Hz \\
Window duration & 30 s \\
Source-grid samples / window & 3,750 \\
Stored samples / window & 1,500 \\
Patch duration & 1 s / 50 samples \\
Maximum candidate windows / case & 200 \\
Maximum configured missing fraction & 0.20 \\
Schema version & 1 \\
\bottomrule
\end{tabular}
\end{table}

\paragraph{Track selection and common temporal grid.}
For each case, the converter selects PPG, ECG, respiratory-related, and
arterial-pressure tracks using ordered candidates followed by
case-insensitive suffix matching. The default candidates are
\code{SNUADC/PLETH} and \code{Solar8000/PLETH} for PPG;
\code{SNUADC/ECG\_II}, \code{SNUADC/ECG\_I}, and
\code{SNUADC/ECG\_III} for ECG;
\code{SNUADC/RESP} and \code{SNUADC/CO2} for respiratory-related
waveforms; and \code{SNUADC/ART}, \code{SNUADC/ABP}, and
\code{Solar8000/ART} for arterial pressure.
Because suffix matching is allowed, the selected device track need not
belong to this default list; for example, the three raw-record
reconstruction cases described below selected \code{Primus/CO2} as the
respiratory-related waveform.

All selected tracks for a case are passed jointly to the VitalDB SDK and
sampled onto one 125-Hz output grid. Candidate windows are non-overlapping
3,750-sample intervals, and the same \code{start:stop} indices are applied
to every selected modality. When more than 200 candidate intervals are
available, at most 200 are selected deterministically across the case
duration using integer-spaced indices with duplicate indices removed.
A PPG-invalid interval is discarded; an invalid optional modality instead
remains a PPG window with the corresponding availability flag unset.

This construction establishes that the conversion code uses a shared SDK
grid and identical window indices across modalities. It does
\emph{not} establish zero acquisition-device offset, absence of clock
drift, or independent correctness of native timestamp decoding.
In particular, ECG and PPG peaks are not forcibly aligned: their temporal
separation can contain genuine physiological propagation delay and is
part of the definition of pulse-arrival-time supervision.

\paragraph{Window processing and stored schema.}
A modality is considered processable when at least 80\% of the original
window samples are finite, at least two finite samples are present, and
the interpolated waveform is not effectively constant. Missing samples
within an accepted window are filled by linear interpolation over sample
indices. PPG is filtered with a second-order 0.5--8~Hz Butterworth
band-pass using forward--backward filtering, resampled to 50~Hz when
needed, and min--max normalized to $[-1,1]$ within the window. Valid ECG
is independently resampled to the same stored length and normalized;
when ECG is unavailable, a zero placeholder is stored together with
\code{has\_ecg=False}. These are offline storage transformations and do
not imply causal preprocessing of the original continuous signal.

Scalar scaffold targets are extracted from the cleaned 125-Hz windows,
rather than recomputed from the normalized 50-Hz stored waveforms.
The stored schema contains
\code{signals}, \code{ecg\_wave}, \code{metrics},
\code{event\_labels}, \code{patch\_sqi},
\code{feat\_amp}, \code{feat\_skew}, \code{feat\_avg},
\code{has\_ecg}, \code{has\_resp}, \code{has\_abp},
\code{subject\_id}, \code{case\_id}, \code{source\_id},
\code{window\_start\_sample}, and \code{dataset\_id}.
The metric vector is ordered as
HR, RMSSD, rhythm, respiratory rate, respiratory phase, SBP, DBP, MAP,
PAT, SQI, and artifact label.

\paragraph{Modality availability and scaffold provenance.}
We scanned all 114 stored shards. Table~\ref{tab:vitaldb-modality-availability}
reports availability over all 1,149,347 windows. Regression targets are
counted as valid when finite; categorical targets additionally require a
nonnegative stored class value. Modality presence and scaffold validity
are deliberately reported separately: a present waveform can fail target
extraction, while selected targets can use a PPG-derived fallback.

When valid ECG is available, HR, RMSSD, and rhythm are extracted from ECG.
If ECG is unavailable, HR and RMSSD may instead be estimated from PPG
peaks, whereas rhythm is not generated by that fallback. Similarly,
respiratory rate may be estimated from PPG when no respiratory-related
waveform is available, whereas respiratory phase requires the
respiratory-related waveform. SBP, DBP, and MAP require valid arterial
pressure. Consequently, a finite scaffold target is not uniformly
evidence of independent cross-modal supervision.

\begin{table}[!htbp]
\caption{Stored modality and scaffold availability over all 1,149,347
multimodal VitalDB windows. Modality presence and target validity are
separate because extraction can fail and selected targets admit PPG
fallbacks. Percentages use all stored windows as the denominator.}
\label{tab:vitaldb-modality-availability}
\centering
\small
\setlength{\tabcolsep}{5pt}
\begin{tabular}{@{}llrr@{}}
\toprule
Category & Stored field / target & Valid windows & Coverage (\%) \\
\midrule
Modality flag & \code{has\_ecg}  & 1,148,820 & 99.95 \\
Modality flag & \code{has\_resp} & 1,036,879 & 90.21 \\
Modality flag & \code{has\_abp}  &   651,138 & 56.65 \\
\midrule
Scaffold target & HR                & 1,141,818 & 99.34 \\
Scaffold target & RMSSD             & 1,141,818 & 99.34 \\
Scaffold target & Rhythm            & 1,141,291 & 99.30 \\
Scaffold target & Respiratory rate  & 1,148,046 & 99.89 \\
Scaffold target & Respiratory phase & 1,036,879 & 90.21 \\
Scaffold target & SBP               &   568,604 & 49.47 \\
Scaffold target & DBP               &   568,604 & 49.47 \\
Scaffold target & MAP               &   568,604 & 49.47 \\
Stored scalar metric & PAT          &   746,388 & 64.94 \\
Scaffold target & SQI               & 1,149,347 & 100.00 \\
Scaffold target & Artifact label    & 1,149,347 & 100.00 \\
\bottomrule
\end{tabular}
\end{table}

The stored scalar PAT count in Table~\ref{tab:vitaldb-modality-availability}
is not the coverage of the future-beat PAT target used by the state
forecasting objective. Future-beat PAT is recomputed by the future-state
extractor and additionally depends on future-beat validity, finite ECG,
and \code{ecg\_available}. We therefore report these two quantities
separately.

\paragraph{Raw-record reconstruction check.}
As a direct raw-to-stored consistency check, we reconstructed the first
stored window from the first, middle, and last sorted shards, corresponding
to source cases 1, 3202, and 6336. For all three examples, the stored PPG
and ECG waveforms, all 11 scalar metrics, patch-level SQI, event labels,
and auxiliary feature arrays were reproduced with zero maximum absolute
difference over finite elements. Finite-value masks and
\code{has\_ecg}/\code{has\_resp}/\code{has\_abp} flags also matched.
A separately written implementation of the window-level PPG and ECG
processing steps reproduced the same stored PPG and ECG arrays.

The scope of this check is intentionally limited. All three audited
windows begin at source-grid index zero; they therefore do not test
nonzero temporal offsets or the full diversity of track selections and
missing-modality patterns. The independently written waveform processing
still uses the same VitalDB SDK temporal grid and the same basic SciPy
operators, and the scalar metric extractors were not independently
reimplemented. We therefore interpret this result as a raw-window
reconstruction and partial processing cross-check, rather than an
exhaustive corpus reconstruction or an independent validation of
acquisition-device synchronization.

\paragraph{Manifest granularity and audit status.}
The recovered corpus-level manifest records the conversion configuration,
case-level completion/failure counts, and global output inventory. Each stored
HDF5 row additionally carries source/case identifiers, the source-grid start
index, and modality flags. These artifacts support row-level traceability to a
source case and window, but they are not yet a standalone exported
one-row-per-window manifest containing the final selected raw track name and
scaffold provenance for every row. We therefore distinguish the checks already
completed---the full 114-shard storage scan and the three-window raw
reconstruction---from the independent audit protocol specified below.

\paragraph{Independent audit protocol.}
We specify the following protocol for a converter-independent audit. First,
export a record/window manifest mapping every stored row to its source case,
selected raw tracks, 125-Hz source-grid start index, output shard/row,
modality flags, and scaffold provenance, and verify a one-to-one
manifest--HDF5 correspondence, including duplicate, missing, and failed cases.
Second, draw a prespecified deterministic stratified sample spanning nonzero
source offsets, early/late case segments, alternative device tracks, missing
modalities, interpolated intervals, and signal-quality strata.

For each sampled window, reconstruct the selected raw tracks and the
native-time-to-125-Hz mapping without calling the converter's window-processing
helpers. Apply the documented window indices and compare arrays, finite-value
masks, modality flags, and scaffold provenance against the stored row under
predeclared tolerances. Native timestamp semantics and device offsets must be
checked directly rather than inferred by forcing ECG--PPG peak alignment.
Finally, verify that every finite scaffold target is compatible with either its
required modality or a documented PPG fallback, retain all failures without
post hoc exclusion, and record hashes of the converter, auditor, manifests, and
audited raw records.

The evidence completed for this submission establishes full-corpus stored
availability and exact reconstruction for three sampled source-grid-start
windows. It does not establish full-corpus raw reconstruction, independently
decoded native timestamps for every device pair, or byte-identical provenance
for every historical shard.

\FloatBarrier

\subsection{Final training and software configuration}
\label{app:training-config}

The canonical optimizer, batch, EMA, hardware, and run-length settings are
listed in Table~\ref{tab:final-protocol}. Optimization uses fused AdamW with
$\beta=(0.9,0.95)$ and weight decay 0.05, a 10-epoch linear warmup from zero to
$6\times10^{-4}$ followed by cosine decay to $10^{-5}$, gradient clipping at
1.0, BF16, and EMA decay 0.9998 with the tokenizer excluded. Each 80-epoch run
uses 2 nodes $\times$ 8 A100 GPUs, per-GPU batch 128, gradient accumulation 2,
and global batch 4096; seeds are $\{0,1,2\}$.

The software environment uses PyTorch 2.6.0+cu118 with CUDA runtime 11.8 and an
NVIDIA driver reporting CUDA 12.2 compatibility, together with
\code{timm==0.9.16}, \code{einops}, \code{h5py}, \code{wfdb},
\code{tensorboardX}, and \code{rotary\_embedding\_torch}. Performance mode is
intentionally nondeterministic. Process seeds initialize as
$s+\mathrm{rank}$ and update each epoch to
$s+e\cdot\mathrm{world}+\mathrm{rank}$ while patient splits remain fixed.

\paragraph{Historical pretraining cost.}
For the full-model seed-0 run, recorded epoch bodies total 58,638.7 seconds
(16.289 hours) across 16 GPUs, corresponding to 260.616 GPU-hours; recorded
mean training throughput is 7,041.03 windows/s. These measurements exclude
dataloader initialization and checkpoint I/O and describe one pretraining run,
not the total study cost. Historical peak training memory is unavailable, so
inference-memory measurements are not used as a proxy.

\FloatBarrier

\subsection{Numerical-stability checks}

\begin{table}[!htbp]
\caption{Numerical-stability check for prefix normalization and the robust
hidden-waveform objective. Each source contributes 40,960 window--cutoff
cases. These values describe objective stability rather than trained-model
accuracy.}
\label{tab:stability-app}
\centering
\small
\begin{tabular*}{0.72\textwidth}{@{\extracolsep{\fill}}lrr@{}}
\toprule
Statistic & MIMIC & VitalDB \\
\midrule
Scale-floor activation (\%) & 0.164 & 0.601 \\
Maximum normalized suffix MSE & 68.84 & 91.40 \\
Maximum robust waveform loss & 9.99 & 10.77 \\
\bottomrule
\end{tabular*}
\end{table}

\section{Forecasting Diagnostics and Boundary Audits}
\label{app:future-eval}

\subsection{State-wise forecasting table}

Table~\ref{tab:future-statewise-app} expands the main intrinsic result by state and horizon. Values are averaged over the five legal cutoffs. PAT is reported conditionally because it requires synchronized ECG and a valid R-peak.

\begin{table}[!htbp]
\caption{State-wise intrinsic forecasting for the final \method{} model.
MAE-Skill is computed from patient-group-macro MAE relative to the
training-median predictor on the same valid support.
For each horizon, values are averaged equally over five fixed cutoffs
and three pretraining seeds.
Spearman correlations are computed over valid elements within each
evaluation cell; correlations and coverage are then averaged over
cutoffs, seeds, and all four horizons.
PAT is evaluated conditionally on ECG availability and valid target
extraction; its coverage denominator includes only ECG-available windows.
Coverage measures target availability, not extraction correctness.}
\label{tab:future-statewise-app}
\centering
\scriptsize
\setlength{\tabcolsep}{2.6pt}
\renewcommand{\arraystretch}{1.06}

\begin{tabular*}{\linewidth}{@{\extracolsep{\fill}}lcccccccc@{}}
\multicolumn{9}{c}{\textit{Forecast skill by source and horizon}}\\[-1.5pt]
\toprule
& \multicolumn{4}{c}{MIMIC skill $\uparrow$}
& \multicolumn{4}{c}{VitalDB skill $\uparrow$} \\
\cmidrule(lr){2-5}\cmidrule(lr){6-9}
State & H1 & H2 & H3 & H4 & H1 & H2 & H3 & H4 \\
\midrule
$\log$ IBI
& 0.5694 & 0.5443 & 0.5396 & 0.5354
& 0.6635 & 0.6505 & 0.6439 & 0.6392 \\
$\log$ amplitude
& 0.3188 & 0.2552 & 0.2207 & 0.1911
& 0.4335 & 0.3659 & 0.3421 & 0.3213 \\
$\log$ rise time
& 0.4203 & 0.3956 & 0.3896 & 0.3852
& 0.3089 & 0.2952 & 0.2921 & 0.2903 \\
$\log$ half-height width
& 0.4555 & 0.4322 & 0.4277 & 0.4262
& 0.5267 & 0.5054 & 0.5005 & 0.4991 \\
$\log$ pulse area
& 0.2924 & 0.2219 & 0.2020 & 0.1858
& 0.4165 & 0.3507 & 0.3303 & 0.3163 \\
Onset baseline
& 0.3209 & 0.2411 & 0.2206 & 0.2018
& 0.4058 & 0.3437 & 0.3230 & 0.3099 \\
Rise/fall asymmetry
& 0.1878 & 0.1523 & 0.1506 & 0.1490
& 0.2996 & 0.2838 & 0.2813 & 0.2800 \\
Fast Haar energy
& 0.4848 & 0.4574 & 0.4515 & 0.4507
& 0.6004 & 0.5815 & 0.5761 & 0.5743 \\
Mid-band Haar energy
& 0.4577 & 0.4293 & 0.4251 & 0.4242
& 0.5401 & 0.5189 & 0.5144 & 0.5139 \\
$\log$ PAT (conditional)
& 0.0585 & 0.0560 & 0.0562 & 0.0538
& 0.1159 & 0.1146 & 0.1141 & 0.1136 \\
\bottomrule
\end{tabular*}

\vspace{4pt}

\begin{tabular*}{0.82\linewidth}{@{\extracolsep{\fill}}lcccc@{}}
\multicolumn{5}{c}{\textit{Association and valid-target coverage}}\\[-1.5pt]
\toprule
& \multicolumn{2}{c}{Spearman $\rho\uparrow$}
& \multicolumn{2}{c}{Coverage (\%)} \\
\cmidrule(lr){2-3}\cmidrule(lr){4-5}
State & MIMIC & VitalDB & MIMIC & VitalDB \\
\midrule
$\log$ IBI
& 0.8081 & 0.8731 & 99.52 & 98.99 \\
$\log$ amplitude
& 0.5538 & 0.6625 & 99.80 & 99.39 \\
$\log$ rise time
& 0.7133 & 0.6411 & 99.80 & 99.39 \\
$\log$ half-height width
& 0.7672 & 0.8132 & 99.80 & 99.39 \\
$\log$ pulse area
& 0.5789 & 0.7070 & 99.80 & 99.39 \\
Onset baseline
& 0.6005 & 0.7025 & 99.80 & 99.39 \\
Rise/fall asymmetry
& 0.5150 & 0.6619 & 99.80 & 99.39 \\
Fast Haar energy
& 0.7646 & 0.8701 & 99.80 & 99.39 \\
Mid-band Haar energy
& 0.7505 & 0.8331 & 99.80 & 99.39 \\
$\log$ PAT (conditional)
& 0.3367 & 0.4600 & 97.42 & 93.56 \\
\bottomrule
\end{tabular*}
\end{table}

\subsection{Horizon, cutoff, and ECG-conditional PAT diagnostics}
\label{app:secondary-forecasting}

\begin{table}[!htbp]
\centering
\caption{Horizon degradation on support shared by all four horizons and both methods, averaged over five cutoffs. Positive $\Delta_{4-1}$ indicates increasing error. \method{} reports mean$\pm$sample SD over three seeds. This stricter support differs from Table~\ref{tab:future-macro}.}
\label{tab:future-horizon-degradation}
\scriptsize
\setlength{\tabcolsep}{3.4pt}
\begin{tabular}{@{}llccccc@{}}
\toprule
Source & Method & $H_1$ MAE & $H_4$ MAE & $\Delta_{4-1}$ &
Relative $\Delta$ (\%) & Slope/beat \\
\midrule
MIMIC & Persistence & 0.13956 & 0.14602 & 0.00646 & 4.63 & 0.00201 \\
MIMIC & \method{} & $0.09755\pm0.00016$ & $0.10669\pm0.00006$ &
$0.00913\pm0.00017$ & $9.37\pm0.18$ & $0.00289\pm0.00005$ \\
\midrule
VitalDB & Persistence & 0.10249 & 0.10654 & 0.00405 & 3.95 & 0.00123 \\
VitalDB & \method{} & $0.07758\pm0.00014$ & $0.08452\pm0.00004$ &
$0.00694\pm0.00018$ & $8.94\pm0.24$ & $0.00224\pm0.00006$ \\
\bottomrule
\end{tabular}
\end{table}

\paragraph{Horizon dependence.}
Table~\ref{tab:future-horizon-degradation} holds support fixed across all four horizons. \method{} retains lower absolute $H_4$ MAE on both sources, but its relative $H_1$--$H_4$ increase exceeds persistence: 9.37\% versus 4.63\% on MIMIC and 8.94\% versus 3.95\% on VitalDB. Its Skill falls from $0.3911\pm0.0011$ to $0.3291\pm0.0004$ and from $0.4672\pm0.0009$ to $0.4207\pm0.0004$, respectively. Thus, lower forecast error through four accepted beats does not imply slower degradation with event horizon.

\paragraph{Cutoff sensitivity.}
Table~\ref{tab:future-cutoff-sensitivity} reports all five fixed context lengths without selecting a preferred cutoff. \method{} outperforms persistence throughout this range. Because moving the cutoff also changes target events and potentially valid support, these comparisons assess sensitivity across forecasting conditions rather than isolate the causal effect of additional context.

\begin{table}[!htbp]
\centering
\caption{Cutoff sensitivity across all five fixed context lengths. Values average four horizons with support paired between methods within each cutoff--horizon cell. \method{} reports mean$\pm$sample SD over three seeds.}
\label{tab:future-cutoff-sensitivity}
\scriptsize
\setlength{\tabcolsep}{3.2pt}
\resizebox{\textwidth}{!}{%
\begin{tabular}{@{}llcccccc@{}}
\toprule
& & \multicolumn{2}{c}{MAE $\downarrow$} &
\multicolumn{2}{c}{MAE Skill $\uparrow$} &
\multicolumn{2}{c}{$\rho\uparrow$} \\
\cmidrule(lr){3-4}\cmidrule(lr){5-6}\cmidrule(lr){7-8}
Source & $t_C$ (s) & Persistence & \method{} & Persistence & \method{} &
Persistence & \method{} \\
\midrule
\multirow{5}{*}{MIMIC}
& 17 & 0.14487 & $0.10439\pm0.00010$ & 0.0895 & $0.3491\pm0.0007$ & 0.4860 & $0.6713\pm0.0004$ \\
& 18 & 0.14450 & $0.10415\pm0.00006$ & 0.0908 & $0.3500\pm0.0004$ & 0.4862 & $0.6724\pm0.0004$ \\
& 19 & 0.14432 & $0.10406\pm0.00008$ & 0.0929 & $0.3509\pm0.0005$ & 0.4865 & $0.6729\pm0.0004$ \\
& 20 & 0.14452 & $0.10383\pm0.00007$ & 0.0900 & $0.3514\pm0.0004$ & 0.4871 & $0.6737\pm0.0004$ \\
& 21 & 0.14474 & $0.10365\pm0.00008$ & 0.0864 & $0.3512\pm0.0005$ & 0.4873 & $0.6738\pm0.0005$ \\
\midrule
\multirow{5}{*}{VitalDB}
& 17 & 0.10724 & $0.08354\pm0.00007$ & 0.2660 & $0.4339\pm0.0005$ & 0.6315 & $0.7512\pm0.0004$ \\
& 18 & 0.10693 & $0.08321\pm0.00006$ & 0.2666 & $0.4348\pm0.0004$ & 0.6326 & $0.7522\pm0.0004$ \\
& 19 & 0.10682 & $0.08307\pm0.00005$ & 0.2667 & $0.4358\pm0.0003$ & 0.6339 & $0.7529\pm0.0002$ \\
& 20 & 0.10675 & $0.08289\pm0.00006$ & 0.2676 & $0.4371\pm0.0005$ & 0.6340 & $0.7538\pm0.0005$ \\
& 21 & 0.10672 & $0.08298\pm0.00008$ & 0.2677 & $0.4361\pm0.0006$ & 0.6345 & $0.7535\pm0.0004$ \\
\bottomrule
\end{tabular}%
}
\end{table}

\paragraph{ECG-conditional PAT.}
PAT labels require synchronized ECG and are excluded from the primary nine-state outcome. \method{} still receives PPG alone. PAT persistence conditions on an observed ECG-derived quantity, so Table~\ref{tab:future-pat} is a conditional diagnostic rather than an input-matched PPG-only comparison.

\begin{table}[!htbp]
\centering
\caption{ECG-conditional PAT diagnostic on paired support, averaged over five cutoffs and four horizons. Skill uses transformed-space MAE; physical MAE uses the inverse transform in milliseconds. PAT persistence conditions on an observed ECG-derived quantity. \method{} receives PPG alone and reports mean$\pm$sample SD over three seeds.}
\label{tab:future-pat}
\scriptsize
\setlength{\tabcolsep}{3.2pt}
\resizebox{\columnwidth}{!}{%
\begin{tabular}{@{}llccccc@{}}
\toprule
Source & Method & Transformed MAE $\downarrow$ & Physical MAE (ms) $\downarrow$ &
Skill $\uparrow$ & $\rho\uparrow$ & Coverage (\%) \\
\midrule
\multirow{3}{*}{MIMIC}
& Train median & 0.37767 & 116.03 & 0.0000 & -- & 94.85 \\
& Persistence & 0.40614 & \textbf{107.88} & -0.0754 & \textbf{0.4013} & 94.85 \\
& \method{} & $\mathbf{0.35708\pm0.00008}$ & $109.14\pm0.05$ &
$\mathbf{0.0545\pm0.0002}$ & $0.3379\pm0.0061$ & 94.85 \\
\midrule
\multirow{3}{*}{VitalDB}
& Train median & 0.42801 & 130.59 & 0.0000 & -- & 89.03 \\
& Persistence & 0.55986 & 153.75 & -0.3080 & 0.2749 & 89.03 \\
& \method{} & $\mathbf{0.37790\pm0.00022}$ & $\mathbf{115.77\pm0.06}$ &
$\mathbf{0.1171\pm0.0005}$ & $\mathbf{0.4664\pm0.0004}$ & 89.03 \\
\bottomrule
\end{tabular}%
}
\end{table}

PAT MAE and Skill improve in transformed space in all 40 source--cutoff--horizon cells, whereas correlation improves in only 20. On MIMIC, lower transformed error coexists with higher physical MAE (109.14 versus 107.88~ms) and lower $\rho$ (0.3379 versus 0.4013). On VitalDB, all four metrics improve. Lower PAT coverage reflects the additional synchronized-ECG and matching requirements. These source- and metric-dependent results do not establish uniform improvement in physical timing accuracy or clinical utility.

\FloatBarrier

\subsection{Stored-suffix intervention audit}

We evaluate the information boundary at the stored-window interface
using 10,000 distinct windows from each of the MIMIC and VitalDB
patient-level test splits, selected by group-balanced sampling.
The same sampled windows are evaluated using checkpoint seeds
$\{0,1,2\}$ at zero-based epoch 79 and cutoffs
$\{17,18,19,20,21\}$~s. Distinct sampled windows are not assumed
to be statistically independent.

For each audited prefix, we replace the stored suffix with
(i) a donor suffix from a different patient in the same source
and quality bin,
(ii) its time reversal,
(iii) a deterministic perturbation matched to its mean and
standard deviation,
(iv) zeros, or
(v) an extreme finite waveform.
Donor quality bins are determined from the mean finite patch-level
signal-quality score over the first 17 patches.
The prefix, cutoff, model parameters, and forward-pass randomness
are held fixed. Evaluation uses deterministic FP32 execution
with TF32 disabled.

The audit checks five output groups: visible PPG states,
visible variate states, typed registers, forecast context,
and the $4\times10$ future-state predictions.
Its 750 invariance summary rows correspond to two sources,
three checkpoint seeds, five cutoffs, five interventions,
and five output groups. Each row summarizes 10,000 windows.
The gradient audit contains 30 summary rows, corresponding
to two sources, three checkpoint seeds, and five cutoffs.
No audited window failed the corresponding acceptance checks.

\begin{table}[!htbp]
\caption{Completed stored-suffix intervention and gradient audit.
Deviations are maximum absolute elementwise differences over
the audited windows, checkpoint seeds, and cutoffs.
The visible-state column takes the maximum over both visible
PPG and visible variate states.
These results apply to the stored-window interface; they do
not establish invariance of upstream raw-signal preprocessing.}
\label{tab:boundary-audit-app}
\centering
\scriptsize
\setlength{\tabcolsep}{2.8pt}
\renewcommand{\arraystretch}{1.08}

\begin{tabular*}{\linewidth}
{@{\extracolsep{\fill}}lcccc@{}}
\toprule
Suffix intervention
& \makecell{Max visible-state\\deviation}
& \makecell{Max register\\deviation}
& \makecell{Max context\\deviation}
& \makecell{Max prediction\\deviation} \\
\midrule
Same-source quality-bin-matched donor & 0 & 0 & 0 & 0 \\
Time reversal                        & 0 & 0 & 0 & 0 \\
Moment-matched deterministic perturbation
                                     & 0 & 0 & 0 & 0 \\
Zeros                                & 0 & 0 & 0 & 0 \\
Extreme finite waveform              & 0 & 0 & 0 & 0 \\
\bottomrule
\end{tabular*}

\vspace{5pt}

\begin{tabular*}{0.88\linewidth}
{@{\extracolsep{\fill}}lr@{}}
\toprule
Gradient / acceptance quantity & Observed result \\
\midrule
Maximum absolute suffix-input gradient & 0 \\
Minimum prefix-gradient $L_2$ norm & 34.95332718 \\
Minimum stabilized prefix-to-suffix ratio
& $3.49533265\times10^{31}$ \\
Invariance failure rows / summary rows & 0 / 750 \\
Gradient failure rows / summary rows & 0 / 30 \\
\bottomrule
\end{tabular*}
\end{table}

For the gradient audit, let $\mathbf{c}(x)$ denote the forecast
context and $\widehat{\mathbf{S}}(x)$ the future-state prediction
matrix for a stored input window $x$. We differentiate the
fixed scalar
\[
q(x)
=
\|\mathbf{c}(x)\|_2^2
+
\|\widehat{\mathbf{S}}(x)\|_F^2.
\]
Writing its input gradient as prefix and suffix components
$g_{\mathrm{pre}}$ and $g_{\mathrm{suf}}$, respectively,
the reported ratio is
\[
R(x)
=
\frac{\|g_{\mathrm{pre}}\|_2}
{\max\!\left(\|g_{\mathrm{suf}}\|_\infty,10^{-30}\right)}.
\]
The gradient acceptance criteria require
$\|g_{\mathrm{suf}}\|_\infty\leq10^{-12}$,
$\|g_{\mathrm{pre}}\|_2>10^{-8}$, and $R(x)\geq10^4$.
The invariance tolerance is the larger of $10^{-6}$ and ten
times the repeated-forward numerical deviation.
The recorded repeated-forward deviation and all intervention
deviations are zero.

All five interventions leave the audited visible states,
typed registers, forecast context, and predictions unchanged
at recorded precision. The nonzero prefix gradient establishes
local input sensitivity of the audited scalar at the tested
inputs, excluding an input-insensitive explanation for these
observations. The large finite ratio follows from the
$10^{-30}$ denominator floor when the suffix gradient is zero;
it is a numerical diagnostic, not a physiological effect size.
These finite-sample intervention and scalar-gradient checks
support the implemented stored-window boundary, but do not
constitute an exhaustive test of all suffix perturbations or
a full-Jacobian proof of invariance.

\FloatBarrier

\subsection{Target-quality audit}

The future states are deterministic physiological proxy targets rather
than clinical gold standards.
We completed an automated audit of 20,000 stored windows, comprising
10,000 windows from MIMIC and 10,000 from VitalDB.
Synchronized ECG was available for 3,785 MIMIC windows and all
10,000 VitalDB windows.
The audit examines cross-signal beat agreement, interval consistency,
reference heart-rate error, and target availability across signal-quality
strata.
These measurements assess the extraction pipeline; they are not
forecasting-performance metrics.

\paragraph{Automated cross-signal agreement.}
Automatically detected ECG peaks are matched one-to-one to PPG peaks
using a positive PPG lag.
The implemented matching interval is 2--32 samples at 50\,Hz
(40--640\,ms), obtained by rounding the nominal 50--650\,ms bounds.
Sensitivity is the number of matched pairs divided by the number of
detected ECG peaks; positive predictive value (PPV) uses the number of
detected PPG peaks as its denominator.
Both quantities pool peak counts over ECG-available windows.
Because matching uses peaks from the entire stored window, these
measurements are not restricted to the four future target beats.

For interval agreement, the audit compares PPG and ECG intervals
between successive matched pairs, computes the mean absolute difference
within each eligible window, and then averages over windows.
These intervals can span missed detections and should not be interpreted
as exclusively adjacent-beat intervals.
The resulting interval MAEs are 82.65\,ms on MIMIC and 73.48\,ms on
VitalDB.
PPG-derived heart rate has MAE 4.57 and 5.27\,bpm, respectively, against
the available stored reference.
The reference ECG peaks are automatically detected rather than manually
annotated, so disagreement can arise from either detector, cross-signal
alignment, or the matching rule.

\begin{table}[!htbp]
\caption{Automated target-quality assessment. Cross-signal sensitivity
and PPV concern whole-window peak matching, not future-beat prediction.
Interval MAE is averaged over eligible windows, whereas sensitivity and
PPV use pooled peak counts. No pooled cross-source error or detection
score is reported.}
\label{tab:target-quality-app}
\centering
\small
\setlength{\tabcolsep}{4pt}
\renewcommand{\arraystretch}{1.08}
\begin{tabular*}{\textwidth}{@{\extracolsep{\fill}}lccc@{}}
\toprule
Quantity & MIMIC & VitalDB & Aggregate \\
\midrule
Audited windows
& 10,000 & 10,000 & 20,000 \\
ECG-available windows
& 3,785 & 10,000 & 13,785 \\
Matched PPG--ECG interval MAE (ms)
& 82.65 & 73.48 & -- \\
PPG-derived HR vs.\ stored reference MAE (bpm)
& 4.57 & 5.27 & -- \\
ECG-reference peak sensitivity (\%)
& 55.52 & 50.71 & -- \\
ECG-reference peak PPV (\%)
& 73.50 & 81.27 & -- \\
Mean PPG-state target availability (\%)
& 99.61 & 99.28 & -- \\
Mean conditional PAT availability (\%)
& 97.10 & 93.17 & -- \\
\bottomrule
\end{tabular*}
\end{table}

\paragraph{Quality-stratified target availability.}
We stratify windows into source-specific quartiles of the mean finite
patch-level signal-quality score.
Each quartile contains 2,500 windows per source; Q1 and Q4 denote the
lowest and highest score quartiles, respectively.
PPG-state availability averages the validity indicators equally over
nine states, five fixed cutoffs, and four horizons.
PAT availability is averaged over cutoffs and horizons using only
ECG-available windows as the denominator.

As shown in Table~\ref{tab:target-quality-strata-app}, mean PPG-state
availability increases from 98.60\% to 99.98\% between Q1 and Q4 on
MIMIC and from 97.17\% to 100.00\% on VitalDB.
The latter value is rounded and does not imply that every target is
valid.
Conditional PAT availability is lower, particularly in VitalDB Q1
(87.03\%).
These are audit-sample availability rates, not the paired-support
coverage rates used for model-versus-persistence comparisons.

\begin{table}[!htbp]
\caption{Signal-quality-stratified target availability in the automated
audit. Each source-specific quartile contains 2,500 windows.
PPG values average nine states, five cutoffs, and four horizons.
PAT denominators include only ECG-available windows.
Availability reflects acceptance by the deterministic extractor and
does not establish annotation correctness.}
\label{tab:target-quality-strata-app}
\centering
\small
\setlength{\tabcolsep}{7pt}
\renewcommand{\arraystretch}{1.08}
\begin{tabular*}{0.85\textwidth}{@{\extracolsep{\fill}}lcccc@{}}
\toprule
& \multicolumn{2}{c}{PPG-state availability (\%)}
& \multicolumn{2}{c}{Conditional PAT availability (\%)} \\
\cmidrule(lr){2-3}\cmidrule(lr){4-5}
Quality stratum & MIMIC & VitalDB & MIMIC & VitalDB \\
\midrule
Q1: lowest score  & 98.60 & 97.17  & 95.03 & 87.03 \\
Q2               & 99.94 & 99.95  & 97.87 & 93.62 \\
Q3               & 99.92 & 99.99  & 97.61 & 95.17 \\
Q4: highest score & 99.98 & 100.00 & 97.95 & 96.84 \\
\bottomrule
\end{tabular*}
\end{table}

\paragraph{Interpretation and remaining validation.}
High target availability coexists with substantially lower
cross-signal peak sensitivity and PPV.
Consequently, extractor acceptance cannot be treated as evidence that
all accepted peaks, onsets, or morphological descriptors are correct.
The automated audit provides interval-consistency and availability
evidence, but does not independently validate every target dimension
or establish clinical accuracy.

\section{Additional Transfer and Efficiency Analyses}
\label{app:additional-analyses}

\subsection{Label efficiency}

Label efficiency is evaluated on BIDMC HR/RR, PPG-BP SBP, WESAD stress, DaLiA activity, and Stanford quality at nominal 1/5/10/25/100\% training-label budgets. The evaluation covers six methods, five label budgets, six tasks, and downstream seeds 42/43/44. Full \method{} and the no-state control use pretraining seed 0; entries report the mean and sample SD across downstream seeds 42/43/44. Pretrained encoders are frozen, whereas the randomly initialized backbone is trainable. Full-label results are reused from the reference runs rather than mixed with the main-table three-pretraining-seed \method{} aggregate. Group-level allocation and minimum-sample constraints can make the realized budget differ from the nominal percentage, and the observed non-monotonic learning curves are retained without smoothing. BIDMC remains an overlap-sensitive benchmark. GPT-PPG is omitted because it is not part of this label-efficiency comparison.

\begin{table}[!htbp]
\caption{Label efficiency for BIDMC heart rate (MAE $\downarrow$). Values are mean$\pm$sample SD across downstream seeds 42/43/44 at each nominal label budget.}
\label{tab:label-efficiency-1}
\centering
\scriptsize
\setlength{\tabcolsep}{3.2pt}
\renewcommand{\arraystretch}{1.08}
\resizebox{\textwidth}{!}{%
\begin{tabular}{@{}lccccc@{}}
\toprule
Method & 1\% & 5\% & 10\% & 25\% & 100\% \\
\midrule
\method{} (full) & 12.3519$\pm$1.5664 & 10.2062$\pm$0.4755 & 5.7168$\pm$0.5292 & 2.4444$\pm$0.4727 & 1.5859$\pm$0.0225 \\
No-state control & 12.6512$\pm$1.5451 & 11.7393$\pm$0.7005 & 7.9118$\pm$0.8629 & 4.3286$\pm$1.1927 & 2.2502$\pm$0.0245 \\
Random initialization (trainable backbone) & 12.2000$\pm$1.3641 & 9.1159$\pm$0.6738 & 6.6818$\pm$0.9687 & 2.8426$\pm$0.2868 & 1.5845$\pm$0.1008 \\
AnyPPG & 12.5275$\pm$1.7645 & 10.0037$\pm$0.1933 & 7.3707$\pm$0.8775 & 6.1154$\pm$0.8784 & 4.9059$\pm$0.5766 \\
PaPaGei-P & 12.6023$\pm$1.4188 & 12.3943$\pm$1.1062 & 8.9256$\pm$0.9239 & 5.7055$\pm$0.2516 & 6.5821$\pm$0.0486 \\
SIGMA-PPG & 12.6334$\pm$1.5372 & 11.5854$\pm$0.5203 & 8.1748$\pm$0.9362 & 5.1786$\pm$0.4960 & 3.7298$\pm$0.0637 \\
\bottomrule
\end{tabular}%
}
\end{table}

\begin{table}[!htbp]
\caption{Label efficiency for BIDMC respiratory rate (MAE $\downarrow$). Values are mean$\pm$sample SD across downstream seeds 42/43/44 at each nominal label budget.}
\label{tab:label-efficiency-2}
\centering
\scriptsize
\setlength{\tabcolsep}{3.2pt}
\renewcommand{\arraystretch}{1.08}
\resizebox{\textwidth}{!}{%
\begin{tabular}{@{}lccccc@{}}
\toprule
Method & 1\% & 5\% & 10\% & 25\% & 100\% \\
\midrule
\method{} (full) & 2.5806$\pm$0.6604 & 2.5698$\pm$0.2768 & 2.3252$\pm$0.0643 & 2.2037$\pm$0.0234 & 2.2619$\pm$0.0292 \\
No-state control & 2.5918$\pm$0.6778 & 2.4475$\pm$0.2325 & 2.2970$\pm$0.0317 & 2.1479$\pm$0.0650 & 2.1233$\pm$0.0150 \\
Random initialization (trainable backbone) & 2.5366$\pm$0.5333 & 2.3326$\pm$0.0666 & 2.3853$\pm$0.1891 & 2.2610$\pm$0.0693 & 2.3229$\pm$0.1214 \\
AnyPPG & 2.6320$\pm$0.7266 & 2.4053$\pm$0.2367 & 2.3581$\pm$0.1179 & 2.2936$\pm$0.1168 & 2.4915$\pm$0.0917 \\
PaPaGei-P & 2.5987$\pm$0.6578 & 2.5207$\pm$0.0801 & 2.4585$\pm$0.0931 & 2.3387$\pm$0.0525 & 2.2741$\pm$0.0442 \\
SIGMA-PPG & 2.5911$\pm$0.6500 & 2.3817$\pm$0.0672 & 2.3840$\pm$0.0859 & 2.2685$\pm$0.0830 & 2.2468$\pm$0.0500 \\
\bottomrule
\end{tabular}%
}
\end{table}

\begin{table}[!htbp]
\caption{Label efficiency for PPG-BP systolic blood pressure (MAE $\downarrow$). Values are mean$\pm$sample SD across downstream seeds 42/43/44 at each nominal label budget.}
\label{tab:label-efficiency-3}
\centering
\scriptsize
\setlength{\tabcolsep}{3.2pt}
\renewcommand{\arraystretch}{1.08}
\resizebox{\textwidth}{!}{%
\begin{tabular}{@{}lccccc@{}}
\toprule
Method & 1\% & 5\% & 10\% & 25\% & 100\% \\
\midrule
\method{} (full) & 18.9945$\pm$1.9037 & 16.5287$\pm$0.2352 & 16.4773$\pm$0.2954 & 16.6867$\pm$0.1477 & 15.0048$\pm$0.1281 \\
No-state control & 19.1810$\pm$1.8943 & 16.0520$\pm$0.1716 & 16.0590$\pm$0.2788 & 15.8941$\pm$0.1310 & 14.9920$\pm$0.1775 \\
Random initialization (trainable backbone) & 17.5925$\pm$0.2417 & 17.1259$\pm$0.1618 & 17.2089$\pm$0.5684 & 17.1493$\pm$0.1221 & 16.1204$\pm$0.4084 \\
AnyPPG & 19.2395$\pm$1.6249 & 17.0720$\pm$0.4896 & 16.8202$\pm$0.0086 & 16.7418$\pm$0.2213 & 16.8517$\pm$0.1128 \\
PaPaGei-P & 18.9927$\pm$1.8266 & 16.4422$\pm$0.1416 & 16.5689$\pm$0.3572 & 16.8079$\pm$0.3217 & 16.4966$\pm$0.0805 \\
SIGMA-PPG & 17.9882$\pm$1.3742 & 17.0255$\pm$0.1840 & 16.9884$\pm$0.1183 & 17.2274$\pm$0.2308 & 16.4501$\pm$0.1864 \\
\bottomrule
\end{tabular}%
}
\end{table}

\begin{table}[!htbp]
\caption{Label efficiency for WESAD stress (AUROC $\uparrow$). Values are mean$\pm$sample SD across downstream seeds 42/43/44 at each nominal label budget.}
\label{tab:label-efficiency-4}
\centering
\scriptsize
\setlength{\tabcolsep}{3.2pt}
\renewcommand{\arraystretch}{1.08}
\resizebox{\textwidth}{!}{%
\begin{tabular}{@{}lccccc@{}}
\toprule
Method & 1\% & 5\% & 10\% & 25\% & 100\% \\
\midrule
\method{} (full) & 0.9331$\pm$0.0071 & 0.9331$\pm$0.0071 & 0.9501$\pm$0.0043 & 0.9541$\pm$0.0021 & 0.9604$\pm$0.0007 \\
No-state control & 0.9121$\pm$0.0084 & 0.9121$\pm$0.0084 & 0.9327$\pm$0.0119 & 0.9403$\pm$0.0100 & 0.9493$\pm$0.0002 \\
Random initialization (trainable backbone) & 0.7973$\pm$0.0232 & 0.7973$\pm$0.0232 & 0.8364$\pm$0.0157 & 0.8445$\pm$0.0041 & 0.9298$\pm$0.0035 \\
AnyPPG & 0.7754$\pm$0.0704 & 0.7754$\pm$0.0704 & 0.9096$\pm$0.0298 & 0.9400$\pm$0.0035 & 0.9175$\pm$0.0033 \\
PaPaGei-P & 0.8800$\pm$0.0188 & 0.8800$\pm$0.0188 & 0.9115$\pm$0.0164 & 0.9154$\pm$0.0101 & 0.8737$\pm$0.0041 \\
SIGMA-PPG & 0.7095$\pm$0.0138 & 0.7095$\pm$0.0138 & 0.7575$\pm$0.0171 & 0.7769$\pm$0.0128 & 0.8391$\pm$0.0051 \\
\bottomrule
\end{tabular}%
}
\end{table}

\begin{table}[!htbp]
\caption{Label efficiency for DaLiA activity (macro-F1 $\uparrow$). Values are mean$\pm$sample SD across downstream seeds 42/43/44 at each nominal label budget.}
\label{tab:label-efficiency-5}
\centering
\scriptsize
\setlength{\tabcolsep}{3.2pt}
\renewcommand{\arraystretch}{1.08}
\resizebox{\textwidth}{!}{%
\begin{tabular}{@{}lccccc@{}}
\toprule
Method & 1\% & 5\% & 10\% & 25\% & 100\% \\
\midrule
\method{} (full) & 0.1706$\pm$0.0146 & 0.1706$\pm$0.0146 & 0.2148$\pm$0.0172 & 0.2265$\pm$0.0261 & 0.2946$\pm$0.0031 \\
No-state control & 0.1664$\pm$0.0185 & 0.1664$\pm$0.0185 & 0.1964$\pm$0.0130 & 0.2122$\pm$0.0109 & 0.2945$\pm$0.0013 \\
Random initialization (trainable backbone) & 0.1684$\pm$0.0139 & 0.1684$\pm$0.0139 & 0.2159$\pm$0.0068 & 0.2357$\pm$0.0178 & 0.2911$\pm$0.0068 \\
AnyPPG & 0.0876$\pm$0.0145 & 0.0876$\pm$0.0145 & 0.1123$\pm$0.0103 & 0.1222$\pm$0.0182 & 0.2489$\pm$0.0046 \\
PaPaGei-P & 0.1411$\pm$0.0079 & 0.1411$\pm$0.0079 & 0.1908$\pm$0.0016 & 0.1998$\pm$0.0060 & 0.2402$\pm$0.0047 \\
SIGMA-PPG & 0.1274$\pm$0.0175 & 0.1274$\pm$0.0175 & 0.1557$\pm$0.0172 & 0.1600$\pm$0.0142 & 0.1894$\pm$0.0052 \\
\bottomrule
\end{tabular}%
}
\end{table}

\begin{table}[!htbp]
\caption{Label efficiency for Stanford signal quality (macro-AUROC $\uparrow$). Values are mean$\pm$sample SD across downstream seeds 42/43/44 at each nominal label budget.}
\label{tab:label-efficiency-6}
\centering
\scriptsize
\setlength{\tabcolsep}{3.2pt}
\renewcommand{\arraystretch}{1.08}
\resizebox{\textwidth}{!}{%
\begin{tabular}{@{}lccccc@{}}
\toprule
Method & 1\% & 5\% & 10\% & 25\% & 100\% \\
\midrule
\method{} (full) & 0.8862$\pm$0.0039 & 0.8865$\pm$0.0037 & 0.8924$\pm$0.0063 & 0.9075$\pm$0.0051 & 0.9285$\pm$0.0011 \\
No-state control & 0.9012$\pm$0.0092 & 0.9010$\pm$0.0095 & 0.9097$\pm$0.0030 & 0.9210$\pm$0.0038 & 0.9374$\pm$0.0015 \\
Random initialization (trainable backbone) & 0.8769$\pm$0.0095 & 0.8762$\pm$0.0087 & 0.8844$\pm$0.0031 & 0.8914$\pm$0.0006 & 0.9043$\pm$0.0092 \\
AnyPPG & 0.8749$\pm$0.0127 & 0.8749$\pm$0.0126 & 0.8822$\pm$0.0110 & 0.8942$\pm$0.0165 & 0.9114$\pm$0.0025 \\
PaPaGei-P & 0.8437$\pm$0.0078 & 0.8436$\pm$0.0078 & 0.8530$\pm$0.0077 & 0.8597$\pm$0.0053 & 0.8223$\pm$0.0024 \\
SIGMA-PPG & 0.8405$\pm$0.0188 & 0.8401$\pm$0.0186 & 0.8506$\pm$0.0157 & 0.8672$\pm$0.0161 & 0.8862$\pm$0.0036 \\
\bottomrule
\end{tabular}%
}
\end{table}

\paragraph{Observed label-efficiency pattern.}
Transfer gains are task dependent. At 25\% nominal labels, full \method{} attains BIDMC HR MAE 2.4444 and WESAD stress AUROC 0.9541, compared with 4.3286 and 0.9403 for the no-state control. Conversely, the no-state control attains lower PPG-BP SBP MAE (15.8941 versus 16.6867) and higher Stanford quality macro-AUROC (0.9210 versus 0.9075). The randomly initialized, trainable backbone exceeds the frozen \method{} encoder on DaLiA activity at 10\% and 25\% labels. Non-monotonic curves are retained as observed, and the downstream-seed sample SDs are not interpreted as pretraining-seed uncertainty.

\FloatBarrier

\subsection{Exploratory structural ablations}

The no-register variant removes all learned registers, and the no-graph-bias variant removes the physiological graph bias. Table~\ref{tab:exploratory-ablation-app} reports a 5\%-data single-seed development screen. These values are comparable only within this screen and are not included in the confirmatory comparisons.

\begin{table}[!htbp]
\caption{Development-only structural screen (5\% data, single seed).
These values are comparable only within this screen and do not enter the
confirmatory full-vs-MIMIC-only analysis. Bold marks the best value among the three
shown conditions for each task.}
\label{tab:exploratory-ablation-app}
\centering
\scriptsize
\setlength{\tabcolsep}{2.2pt}
\renewcommand{\arraystretch}{1.08}
\begin{tabular*}{\linewidth}{@{\extracolsep{\fill}}lcccccc@{}}
\toprule
Condition
& \makecell{HR\\MAE$\downarrow$}
& \makecell{RR\\MAE$\downarrow$}
& \makecell{SBP\\MAE$\downarrow$}
& \makecell{Stress\\AUC$\uparrow$}
& \makecell{DaLiA\\score$\uparrow$}
& \makecell{Stanford\\AUC$\uparrow$} \\
\midrule
Full \method{}
& 2.929 & 2.115 & \textbf{14.318} & 0.950 & 0.302 & 0.926 \\
Without learned registers
& \textbf{1.107} & 2.137 & 15.517 & 0.950 & 0.282 & \textbf{0.931} \\
Without physiological graph bias
& 2.703 & \textbf{2.056} & 14.650 & \textbf{0.958} & \textbf{0.314} & 0.928 \\
\bottomrule
\end{tabular*}
\end{table}

\FloatBarrier

\subsection{Compute-matched source contribution}

The contribution of VitalDB is assessed only through full \method{} versus the MIMIC-only control under matched total sampled windows, optimizer updates, checkpoint index, model capacity, learning-rate schedule, tokenizer, pretraining seeds, and downstream hyperparameters. The corresponding forecasting and transfer contrasts are reported in Table~\ref{tab:ablation-main}; unmatched development runs are not used as evidence for a VitalDB effect.

\FloatBarrier

\subsection{Efficiency}

We measure encoder inference efficiency under a common
NVIDIA A100-SXM4-80GB environment using PyTorch 2.6.0,
CUDA 11.8, and BF16 eager execution.
Each measurement uses 100 warm-up iterations followed by
1,000 measured iterations.
Inputs contain 1,500 samples per window, and the frozen
representation dimension is 768 for all conditions.
Table~\ref{tab:efficiency-app} reports total parameters,
batch-64 inference throughput, peak allocated GPU memory,
and mean encoder latency at batch sizes 1 and 64.

These are architecture-only measurements without pretrained
weights, rather than end-to-end training benchmarks.
Consequently, the no-register and no-graph-bias measurements
characterize architecture-only computational variants and do not imply
confirmatory pretraining or downstream evaluation for those variants.
Total parameter counts and memory are reported directly in
Table~\ref{tab:efficiency-app}; all memory values use decimal GB.

\begin{table}[!htbp]
\caption{Encoder inference efficiency on an A100-SXM4-80GB.
Throughput and peak allocated memory are measured at batch size 64.
B1 and B64 denote mean latency per batch at batch sizes 1 and 64,
respectively.
Measurements use BF16 eager execution, 100 warm-up iterations,
and 1,000 measured iterations.
Parameter counts include all parameters in the instantiated model;
the timed operation is encoder inference.}
\label{tab:efficiency-app}
\centering
\scriptsize
\setlength{\tabcolsep}{2.5pt}
\renewcommand{\arraystretch}{1.06}

\begin{tabular*}{\linewidth}{@{\extracolsep{\fill}}lrrrrr@{}}
\toprule
Condition
& \makecell{Params\\(M)}
& \makecell{B64 inference\\windows/s}
& \makecell{B64 peak allocated\\memory (GB)}
& \makecell{B1 latency\\(ms)}
& \makecell{B64 latency\\(ms)} \\
\midrule
Full \method{}
& 98.489 & 1549.59 & 1.286 & 14.370 & 41.301 \\
No-state
& 98.489 & 1557.13 & 1.286 & 14.385 & 41.101 \\
Token-matched raw-view
& 98.489 & 1586.05 & 1.286 & 13.842 & 40.352 \\
No registers
& 98.463 & 1835.01 & 1.286 & 14.098 & 34.877 \\
No graph bias
& 98.471 & 1603.16 & 1.286 & 13.739 & 39.921 \\
\bottomrule
\end{tabular*}
\end{table}

Removing registers yields the highest measured batch-64 throughput
and the lowest batch-64 latency among the evaluated architectures.
However, batch-1 latency does not follow the same ordering.
The small differences among several conditions are descriptive:
repeated-run uncertainty was not measured, so these values do not
establish statistically reliable speed differences.
Likewise, lower inference cost alone does not establish a favorable
accuracy--efficiency trade-off for the no-register or no-graph-bias variants.

\end{document}